\documentclass{article}
\usepackage{blindtext}
\usepackage[a4paper, total={4in, 8in}]{geometry}
\usepackage{bbding,fontawesome,pifont}
\usepackage{blindtext}
\usepackage{amsmath}
\usepackage{algorithmicx}
\usepackage[ruled, noend, noline]{algorithm2e}
\usepackage{multirow}
\usepackage{float}
\usepackage[usenames,dvipsnames]{color}
\usepackage{amsfonts}
\usepackage{graphicx,subcaption} 
\usepackage{multirow}
\usepackage{longtable}
\usepackage{tabularx}
\usepackage{arydshln}
\usepackage{array}
\usepackage{adjustbox}
\newcolumntype{L}[1]{>{\raggedright\let\newline\\\arraybackslash\hspace{0pt}}m{#1}}
\newcolumntype{C}[1]{>{\centering\let\newline\\\arraybackslash\hspace{0pt}}m{#1}}
\newcolumntype{R}[1]{>{\raggedleft\let\newline\\\arraybackslash\hspace{0pt}}m{#1}}
\usepackage{dstyle}

\usepackage{lastpage}

\begin{document}

\author{\name Mahmoud Tahmasebi \email mahmoud.tahmasebi@research.atu.ie \\
       \addr Center for Mathematical Modelling and Intelligent Systems for Health and Environment (MISHE)\\
       Atlantic Technological University\\
       Sligo, Ireland
       \AND
       \name Saif Huq \email shuq@bridgewater.edu \\
       \addr Department of Engineering and Physics\\
       Bridgewater College\\
       Bridgewater, USA
       \AND
       \name Kevin Meehan \email kevin.meehan@atu.ie \\
       \addr Center for Mathematical Modelling and Intelligent Systems for Health and Environment (MISHE)\\
        Atlantic Technological University\\
        Donegal, Ireland
       \AND
       \name Marion McAfee \email Marion.McAfee@atu.ie \\
       \addr Center for Mathematical Modelling and Intelligent Systems for Health and Environment (MISHE) \\
       Atlantic Technological University    \\
       Sligo, Ireland}
       
\title{ESMStereo: Enhanced ShuffleMixer Disparity Upsampling for Real-Time and Accurate Stereo Matching}

\maketitle       

\begin{abstract}%

Stereo matching has become an increasingly important component of modern autonomous systems. Developing deep learning-based stereo matching models that deliver high accuracy while operating in real-time continues to be a major challenge in computer vision. In the domain of cost-volume-based stereo matching, accurate disparity estimation depends heavily on large-scale cost volumes. However, such large volumes store substantial redundant information and also require computationally intensive aggregation units for processing and regression, making real-time performance unattainable. Conversely, small-scale cost volumes followed by lightweight aggregation units provide a promising route for real-time performance, but lack sufficient information to ensure highly accurate disparity estimation. To address this challenge, we propose the Enhanced Shuffle Mixer (ESM) to mitigate information loss associated with small-scale cost volumes. ESM restores critical details by integrating primary features into the disparity upsampling unit. It quickly extracts features from the initial disparity estimation and fuses them with image features. These features are mixed by shuffling and layer splitting then refined through a compact feature-guided hourglass network to recover more detailed scene geometry. The ESM focuses on local contextual connectivity with a large receptive field and low computational cost,  leading to improved disparity estimation accuracy while maintaining real-time performance under the evaluated settings. The compact version of ESMStereo achieves an inference speed of 116 FPS on RTX 4070S and 91 FPS on the AGX Orin. The source code is available at \url{https://github.com/M2219/ESMStereo}.

\end{abstract}

\section{Introduction}
\par

\begin{figure}
    \begin{center}
    \includegraphics[scale=0.4]{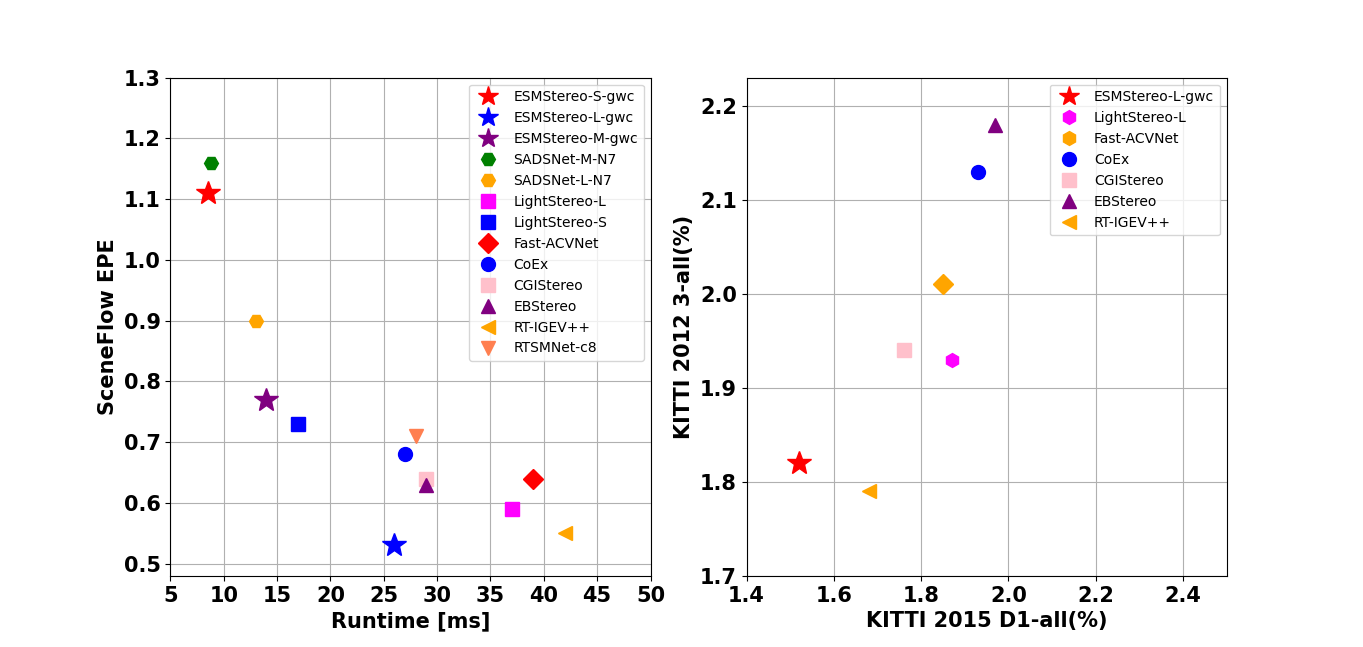}
\caption{Comparison of different ESMStereo variants with state-of-the-art real-time stereo matching methods. The left plot presents SceneFlow EPE versus runtime [ms], while the right plot compares accuracy on the KITTI 2012 \cite{kitti_12} and KITTI 2015 \cite{kitti_15} benchmarks.}
    \label{fig1}
\end{center}
\end{figure}
{\setlength{\parindent}{0cm}
\vspace{5mm}
\par

With the rapid advancement of autonomous systems and robotics in various aspects of life, there is a critical need for real-time and efficient scene perception systems. These systems must process data instantaneously to facilitate real-time decision making, ensure safety, and navigate complex environments effectively whilst also operating within feasible computing power constraints.

{\setlength{\parindent}{0cm}
\vspace{5mm}
\par
Similar to other deep learning tasks such as object detection and segmentation, stereo matching networks play a vital role in the perception block, enabling systems to achieve smooth, human-like depth understanding. However, stereo matching is a computationally intensive task in computer vision. Consequently, much of the research in this domain has focused on balancing the trade-off between speed and accuracy to deliver acceptable performance for computation-constrained applications \cite{xu2024igev++, 26_mobilestereonet, 12_coex, 35_rtsmnet, TAHMASEBI2024129002}.

{\setlength{\parindent}{0cm}
\vspace{5mm}
\par

In cost-volume-based stereo matching, lightweight feature extractors such as MobileNetV2 \cite{saikia2019autodispnet}, utilized in \cite{25_acvnet, xu2024igev++}, and EfficientNetV2 \cite{tan2021efficientnetv2}, adopted in \cite{chen2024mocha, xu2023iterative}, help reduce runtime. However, the primary computational bottleneck lies in the cost volume and the subsequent 3D aggregation block. Constructing a large cost volume across the full range of disparities not only stores significant redundant information \cite{zheng2023diffuvolume} but also necessitates a large 3D convolution-based aggregation unit for processing. On the other hand, naively constructing a small-scale cost volume to achieve real-time performance often results in a significant loss of accuracy.

{\setlength{\parindent}{0cm}
\vspace{5mm}
\par

A small-scale cost volume lacks the capacity to store sufficient matching information, making it challenging to achieve high accuracy. To address this limitation while maintaining real-time performance, various strategies have been proposed. These include employing edge-aware upsampling \cite{1_stereonet}, guiding aggregation with initial features \cite{11_cgistereo, xu2024igev++, 12_coex, electronics14050892}, fusing image features with multi-scale cost volumes \cite{sym17081214}, upsampling low-resolution cost volumes within a bilateral grid guided by image features \cite{9_bgnet, 20_ebstereo}, enriching small-scale cost volumes using a lightweight 2D disparity estimator network \cite{chen2021multi}, and leveraging lightweight feature-guided refinement techniques \cite{18_deeppruner}. These methods aim to mitigate the information loss associated with small-scale cost volumes and enhance real-time stereo matching performance.

{\setlength{\parindent}{0cm}
\vspace{5mm}
\par

Using a single small-scale cost volume and compensating for accuracy loss by informing subsequent modules has shown great potential for achieving real-time stereo matching. However, some studies suggest that utilizing multiple small-scale cost volumes which are appropriately fused can also deliver high accuracy while maintaining high speed performance. For instance, AANet \cite{28_aanet} constructs compact cost volumes enriched with geometric information using multi-scale features, which are subsequently processed through six fused Adaptive Aggregation Modules (AAModules). Similarly, \cite{TAHMASEBI2024129002} introduces a double-cost volume architecture along with a coupling module to effectively fuse parallel aggregation units, demonstrating a robust approach to high speed and accurate stereo matching.

{\setlength{\parindent}{0cm}
\vspace{5mm}
\par

Recent studies such as IINet \cite{li2024iinet}, reporting an EPE of 0.54 px with 26 ms inference time, and LightStereo \cite{guo2024lightstereo}, reporting an EPE of 0.51 px with 54 ms inference time, have shown that pairing a 3D cost volume with a 2D aggregation block can achieve highly accurate disparity estimation in real-time. Replacing 3D operations with 2D ones significantly boosts computational speed. 3D aggregation keeps the disparity dimension along with spatial dimensions, which enables the extraction of rich spatial and geometric information. On the other hand, 2D aggregation processes a 3D cost volume compressed along the disparity range. This compression, however, results in a loss of geometric detail. To address this, such methods often incorporate attention mechanisms and fusion strategies to recover the lost information by leveraging features from the initial input, ensuring accurate and efficient performance.

{\setlength{\parindent}{0cm}
\vspace{5mm}
\par

Typically, a small-scale cost volume results in a low-resolution disparity estimation that requires upsampling to full resolution, which makes the upsampling process crucial for recovering fine details. Recent works, such as \cite{wu2023towards, 35_rtsmnet, 9459444}, have demonstrated that feature-guided upsamplers enable the use of lightweight feature extraction backbones, compact aggregation networks, and small-scale cost volumes while maintaining real-time and accurate performance.

{\setlength{\parindent}{0cm}
\vspace{5mm}
\par

Building on recent advancements, we propose Enhanced ShuffleMixer for real-time Stereo Matching (ESMStereo), as illustrated in Fig.\ref{fig2}. ESMStereo consists of a small-scale cost volume, a lightweight aggregation unit, and an efficient upsampler called Enhanced ShuffleMixer (ESM). As the main contribution, the ESM module processes the initial disparity map by rapidly extracting its features and fusing them with features from the left image. An enhanced ShuffleMixer is employed to mix the fused features through shuffling and layer splitting, focusing on local correlations within the features. The ESM module serves as a unified upsampling component by upsampling the mixed features using a pixel shuffle operation and fusing the disparity map with image features. Pixel shuffling is computationally efficient as it only rearranges existing features and helps to preserve their integrity. However, unlike interpolation, pixel shuffle alone does not inherently smooth or refine the upsampled feature map; hence, a compact feature-guided hourglass network is incorporated. Because of these characteristics, the ESM module plays the main role in reducing inference runtime by enabling the use of a smaller feature extractor and a reduced cost volume.

{\setlength{\parindent}{0cm}
\vspace{5mm}
\par

The whole network ensures smooth disparity estimation while recovering additional scene geometry details. This design allows the network to integrate both global and local contexts, effectively addressing the information loss associated with the small-scale cost volume and the lightweight aggregation block, which results in achieving highly accurate disparity maps in real-time.

\vspace{5mm}
\par

Our main contributions are:

\begin{itemize}
  \item We introduce the Enhanced ShuffleMixer (ESM) module as our primary contribution, a fast feature-guided upsampler designed for efficient upsampling. The ESM module integrates feature fusion, leverages the ShuffleMixer layer to focus on local correlations in the features, and employs a lightweight feature-guided hourglass refinement network to recover additional scene details at each upsampling stage.
  \item The ESMStereo is a stereo matching architecture with low computational complexity designed to  minimize the computational burden of the cost volume and 3D aggregation modules by transferring the computational workload to the post-disparity regression stage, where 2D modules are used to recover the details of the stereo scene.
 \item ESMStereo achieves D1 error rates of 5.4\% on KITTI 2012 and 5.5\% on KITTI 2015, which indicate competitive performance on real-world benchmarks and demonstrate reasonable generalization to unseen datasets compared with other published real-time stereo matching methods.
 
\end{itemize}

\begin{figure}
    \begin{center}
    \includegraphics[scale=0.25]{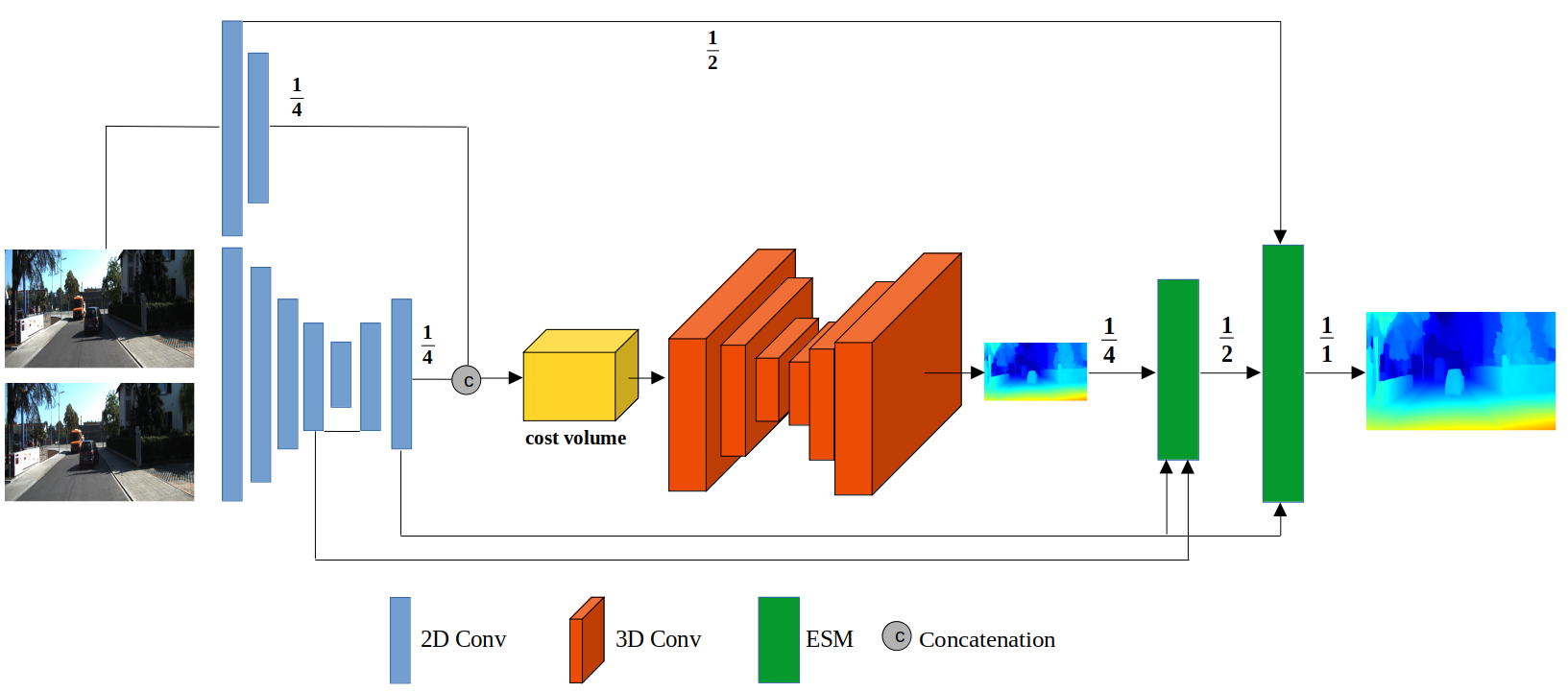}
    \caption{In ESMStereo, the volume is constructed at \(\frac{1}{4}\) resolution and aggregated by a lightweight 3D hourglass network to generate a low resolution initial disparity map which is further up-sampled by ESM modules to estimate the full resolution disparity map. This figure illustrates the architecture for the large version of ESMStereo}
    \label{fig2}
\end{center}
\end{figure}

\vspace{5mm}
\par

\section{Related work}
\subsection{Real-Time Stereo Matching}

In cost volume-based stereo matching, achieving real-time performance largely depends on the construction of the cost volume and the design of the aggregation block, which often relies on computationally intensive 3D convolutional layers. A common approach to improve efficiency is to create a low resolution or sparse cost volume to speed up the network while mitigating accuracy loss by guiding the aggregation block with contextual features \cite{11_cgistereo}, edge information \cite{song2020edgestereo}, or semantic cues \cite{5_rts2net, yang2018segstereo}. Another widely adopted strategy is to shift the computational burden to the post-disparity regression stage, enabling the use of efficient 2D refinement and upsampling modules \cite{35_rtsmnet, 1_stereonet}.
{\setlength{\parindent}{0cm}
\vspace{5mm}
\par

Although feature-guided aggregation blocks, as utilized by CGI-Stereo \cite{11_cgistereo} and CoEx \cite{12_coex}, can help compensate for the limitations of an under-informed small-scale cost volume, the 3D convolutional operations required for aggregation can significantly impact the network's speed. This slowdown occurs because concatenating contextual features with the aggregation layers increases the computational demands on 3D kernels. On the other hand, methods like StereoNet \cite{1_stereonet} and \cite{liu2022convmlp} employ lightweight aggregation blocks and introduce edge-aware refinement modules by concatenating the left image with the initial disparity to recover lost information. While these approaches achieve very high-speed performance, the straightforward use of the left image in the refinement process does not effectively help to recover high-frequency details.

{\setlength{\parindent}{0cm}
\vspace{5mm}
\par

A low-resolution cost volume not only lacks sufficient matching information but also contains redundant data. One approach to enhance accuracy while maintaining a high-speed network is to prune less important information and replace it with richer matching data, as demonstrated by Fast-ACVNet \cite{xu2023accurate}. However, even a small-scale 4D cost volume holds a considerable number of parameters, making it challenging to efficiently identify and eliminate redundant information. Consequently, strategies with low computational complexity tend to limit accuracy, whereas more computationally intensive methods, such as those proposed by ACVNet \cite{25_acvnet} and DiffuVolume \cite{zheng2023diffuvolume}, significantly increase inference time.

{\setlength{\parindent}{0cm}
\vspace{5mm}
\par

Some methods adopt a coarse-to-fine strategy to achieve real-time performance. For example, approaches such as AnyNet \cite{2_anynet}, ADCPNet \cite{8_adcpnet}, GWDRNet \cite{liang2023real}, SADSNet \cite{wu2023towards}, and HRSNet \cite{huang2023real} construct a low-resolution cost volume to estimate an initial disparity map, which is subsequently upsampled and refined at finer levels in a cascading manner. At each level, a new larger cost volume is generated which progressively increases the computational effort. To maintain high speed, these methods typically rely on lightweight feature extraction networks, which adversely affect the quality of the cost volumes and lead to a degradation in accuracy.
\setlength{\parindent}{0pt}
\vspace{5mm}
\par
IINet \cite{li2024iinet} and LightStereo \cite{guo2024lightstereo} are two recent high-performance, real-time methods that utilize 3D cost volumes to reduce computational overhead while employing 2D convolutional layers for aggregation. Since a standalone 2D aggregation block can lead to the loss of structural information, these methods incorporate fusion strategies that combine the aggregation block with initial features and leverage attention mechanisms to recover accuracy. Unlike 2D aggregation such as those used in IINet and LightStereo, 3D aggregation preserves the disparity dimension alongside spatial dimensions, enabling the extraction of rich spatial and geometric information. However, 2D aggregation operates on a 3D cost volume compressed along the disparity range. This means that the 3D cost volume stores much less matching information, which inherently leads to a loss of geometric detail and limits the generalization capability of the network. Therefore, an important direction for improving lightweight stereo matching networks is to restore the lost structural and geometric information during the disparity refinement and upsampling stages.

\vspace{5mm}
\par

\section{Disparity Upsampling}
The most straightforward approach to achieving real-time performance is to use a low resolution cost volume. Typically, this involves pairing the cost volume with an aggregation block to generate a low resolution disparity map, which must then be upsampled to full resolution. As a result, designing an efficient upsampling method presents a significant opportunity to restore scene details while maintaining both accuracy and speed.
\setlength{\parindent}{0pt}
\vspace{5mm}
\par
DispNet \cite{32_dispnet} and RTSNet \cite{6_rtsnet} use deconvolution operations to upsample the disparity map. PSMNet \cite{chang2018pyramid} and StereoNet \cite{1_stereonet} adopt bilinear upsampling, while GwcNet \cite{GWC} and ACVNet \cite{25_acvnet} utilize trilinear upsampling. Although simply using linear upsamplers and deconvolution can quickly produce the full-resolution disparity map, they often fail to capture fine and thin structures. To address this, CoEx \cite{12_coex} proposes superpixel upsampling, a deconvolution-based method guided by initial feature information, which significantly increases accuracy at a low computational cost.
\setlength{\parindent}{0pt}
\vspace{5mm}
\par
Further, CGIStereo \cite{11_cgistereo} and IGEVStereo \cite{xu2023iterative} modify the CoEx \cite{12_coex} upsampling approach by replacing deconvolution operations with nearest-neighbor interpolation. This method refines the disparity map by calculating finer disparities as a weighted sum of the coarse disparities in a \(k \times k\) square neighborhood (e.g., k = 3 or 5), with weights learned from features in the left image. This modification incorporates spatial information whilst also accelerating the upsampling process. 
\setlength{\parindent}{0pt}
\vspace{5mm}
\par

In addition to the methods above, which focus on upsampling disparity estimations in 2D space, some works like BGNet \cite{9_bgnet} and SAGU-Net \cite{wu2023real} attempt to directly upsample the 3D cost volume. BGNet \cite{9_bgnet} and EBStereo \cite{20_ebstereo} present an edge-preserving upsampling module based on the slicing operation within a learned bilateral grid, guided by initial feature maps. SAGU-Net \cite{wu2023real} divides the upsampling computational load between 3D and 2D processes. It introduces a Spatial Attention-Guided Cost Volume Upsampling (SAG-CVU) module to upsample the low-resolution cost volume, followed by a Spatial Attention-Guided Disparity Map Upsampling (SAG-DMU) module to achieve full resolution in 2D space.
\setlength{\parindent}{0pt}
\vspace{5mm}
\par

In the domain of real-time stereo matching systems, SADSNet \cite{wu2023towards} employs deformable convolution \cite{zhu2019deformable} to design the Spatial Adaptive Disparity Shuffle (SADS) upsampler, which calculates each upsampled disparity from coarse disparities within neighboring homogeneous regions. RTSMNet \cite{35_rtsmnet} uses a feature-guided upsampling module to gradually upsample using a parameter-free upsampler based on adaptive kernels \cite{wang2019carafe} and pixel shuffle \cite{shi2016real}. Pixel shuffle is computationally efficient due to its rearrangement-based approach, but unlike interpolation, it does not inherently smooth or refine the upsampled feature map. Thus, using pixel shuffle in a parameter-free upsampler without an effective refinement stage can limit the accuracy of disparity estimation. Building on these methods, FBPGNet \cite{wen2022feature} incorporates a feature subtraction module to capture high-frequency information. These extracted features are upsampled via a deconvolution operation and subsequently used to refine the bilinearly upsampled initial disparity. However, despite using a learnable upsampler, FBPGNet achieves lower accuracy compared to RTSMNet. Since the features tend to lose crucial information during upsampling, this accuracy degradation likely stems from the absence of an efficient refinement process after the deconvolution stage.

\setlength{\parindent}{0pt}
\vspace{5mm}
\par

To overcome the limitations of these approaches, we propose ESMStereo, which incorporates an Enhanced ShuffleMixer (ESM) upsampler followed by an efficient feature-guided refinement module to enhance the bilinearly upsampled initial disparity map. ESMStereo achieves real-time performance while significantly surpassing the accuracy of SADSNet, RTSMNet, and FBPGNet by a substantial margin.

\section{Method}
As shown in Fig.\ref{fig2}, ESMStereo takes two stereo images as input and extracts contextual features using an encoder-decoder architecture. The extracted features, at resolutions of \(\frac{1}{4}\), \(\frac{1}{8}\), and \(\frac{1}{16}\) of the original images, are used to construct compact normalized correlation cost volumes for ESMStereo-L, ESMStereo-M and ESMStereo-S. The constructed cost volume is processed by a lightweight 3D hourglass network, and the aggregated information is regressed to produce a low-resolution disparity map. The disparity map is subsequently upsampled in multiple stages to full resolution using the Enhanced ShuffleMixer (ESM) module. This section introduces the architecture of the ESM module in Sec.\ref{sec:esm}, followed by the feature extraction and cost volume construction in Sec.\ref{sec:feature}. Lastly, the disparity regression process and the loss function used to train the network are detailed in Secs.\ref{sec:reg} and \ref{sec:loss}.

\subsection{Enhanced ShuffleMixer (ESM) Module}
\label{sec:esm}

\begin{figure}
    \begin{center}
    \includegraphics[scale=0.3]{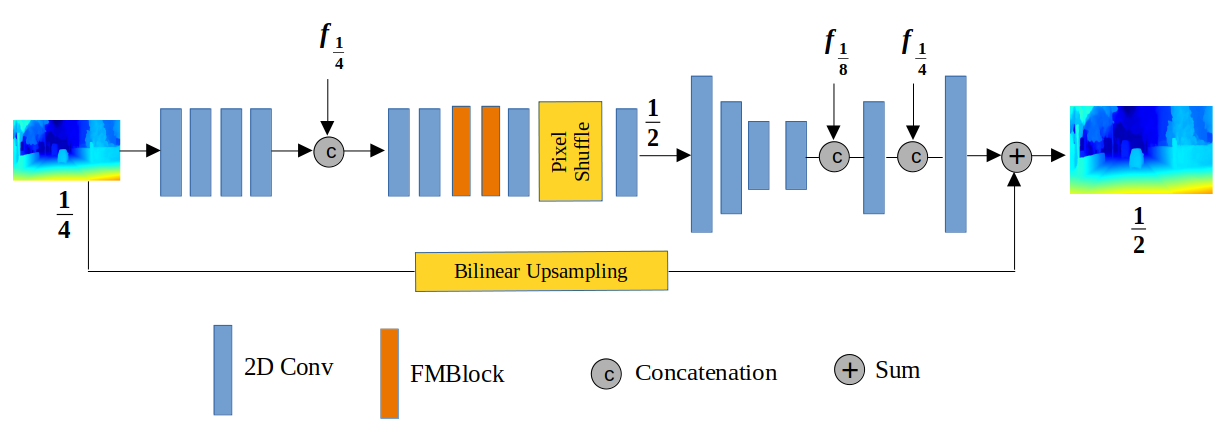}
    \caption{The ESM module extracts features from the low-resolution disparity map using four 2D convolutional layers, which are subsequently fused with contextual features from the left image. The fused features are processed through two FMBlocks \cite{sun2022shufflemixer}, which mix the features by shuffling and splitting the layers to enhance the focus on local correlations within the features. Following this, a feature-guided hourglass refinement network is employed to recover fine structures and produce a smooth disparity estimation.}
    \label{figESM}
\end{center}
\end{figure}

\vspace{5mm}
\par

Using a low-resolution cost volume and a lightweight aggregation unit can result in accuracy degradation due to the insufficient matching information stored in the cost volume and the limited capability of the aggregation unit to produce high-quality, smooth disparity maps. However, the advantage of employing a compact cost volume lies in its ability to significantly improve network speed. To overcome the accuracy limitations, we propose the Enhanced ShuffleMixer (ESM) module. As illustrated in Fig.\ref{figESM}, the ESM module is composed of three main components: a feature fusion block, an enhanced ShuffleMixer upsampler, and a feature-guided refinement network. Together, these elements address the shortcomings of compact cost volumes, enabling the network to achieve both high accuracy and real-time performance.

{\setlength{\parindent}{0cm}
\vspace{5mm}
\par
The first component of the ESM module is a straightforward feature fusion unit. This unit extracts features from the initial disparity estimations using four convolutional layers, which are then concatenated with the corresponding image features. The combined features are further processed through two additional convolutional layers before being passed to the enhanced ShuffleMixer. 

{\setlength{\parindent}{0cm}
\vspace{5mm}
\par

The fusion of multi-level image features provides richer contextual and structural cues, which help the network better preserve edges and fine-grained details during disparity refinement. In addition, incorporating image features stabilizes training by providing strong appearance guidance, reducing ambiguity in textureless regions, and improving convergence behavior. It also enables the model to capture both early geometric structures and high-level semantic consistency, which leads to more accurate and robust disparity estimation.

{\setlength{\parindent}{0cm}
\vspace{5mm}
\par
The enhanced ShuffleMixer utilizes two simplified FMBlocks, as originally proposed in \cite{sun2022shufflemixer}. Each FMBlock consists of two Shuffle Mixing Layers followed by a lightweight convolutional refinement module. 
Each Shuffle Mixing Layer applies LayerNorm, a channel-wise split-point MLP with residual connection, and a depthwise $k \times k$ convolution for efficient local spatial aggregation. 
The final refinement stage uses a bottleneck design with a $3 \times 3$ convolution expanding channels to $C+16$, SiLU activation, and a $1 \times 1$ projection back to $C$, with residual connections throughout to stabilize training. 
FMBlocks perform channel mixing and spatial aggregation through shuffling (channel interaction) and splitting (feature partitioning) in an efficient manner.

{\setlength{\parindent}{0cm}
\vspace{5mm}
\par

After mixing, the features are upsampled using a pixel shuffle operation. Pixel shuffle is computationally efficient since it only rearranges pixel values across channels, which preserves the integrity of the existing features without introducing new information. However, it does not inherently smooth or refine the upsampled features. To overcome this limitation, a feature-guided hourglass refinement network is employed. This network refines the upsampled features by recovering additional details and enriching the scene geometry information. Furthermore, the encoder-decoder structure of the refinement network expands the receptive field, enabling the model to integrate global contextual information effectively. This combination ensures that the ESM module achieves accurate and detailed disparity estimation while maintaining computational efficiency. 

{\setlength{\parindent}{0cm}
\vspace{5mm}
\par

Table \ref{tab:esm_config} summarizes the detailed configuration of the proposed ESM module for the ESMStereo-S, ESMStereo-M, and ESMStereo-L variants. The table specifies the number of ESM stages, input feature dimensions, FMBlock configurations, pixel-shuffle operations, and refinement stages, providing the information required to reproduce the disparity upsampling network.
{\setlength{\parindent}{0cm}
\vspace{5mm}
\par

The three variants follow the same overall design principle of alternating feature mixing, refinement, and pixel-shuffle upsampling stages while differing in depth and upsampling strategy. In ESMStereo-L, the ESM module operates on 1/4-resolution features and performs two successive 2$\times$ pixel-shuffle upsampling stages, progressively reconstructing the disparity map from $1/4 \rightarrow 1/2 \rightarrow 1/1$. ESMStereo-M adopts a deeper design with three ESM stages and three 2$\times$ upsampling operations, enabling gradual disparity refinement from $1/8 \rightarrow 1/4 \rightarrow 1/2 \rightarrow 1/1$. To prioritize computational efficiency, ESMStereo-S employs only two ESM stages and utilizes a larger 4$\times$ pixel-shuffle factor, allowing the disparity representation to be reconstructed from $1/16 \rightarrow 1/4 \rightarrow 1/1$ with fewer intermediate computations.

\begin{table}[t]
\centering
\caption{Detailed configuration of the Enhanced ShuffleMixer (ESM) module for S/M/L variants.}
\label{tab:esm_config}
\begin{tabular}{lccc}
\hline
\textbf{Component} & \textbf{ESMStereo-L} & \textbf{ESMStereo-M} & \textbf{ESMStereo-S} \\
\hline
Number of ESM module & 2 & 3 & 2 \\
Input feature channels & 16 & 8 & 8 \\
FMBlocks count & 2 & 3 & 4 \\
FMBlock kernel sizes & $7 \times 7$ & $7 \times 7$ & $7 \times 7$ \\
Pixel Shuffle units & 2 & 3 & 2 \\
Pixel shuffle upscale factors & 2 & 2 & 4 \\
Upsampling stages & 2 & 3& 2 \\
refinement stages & 2 & 3& 2 \\

\hline
\end{tabular}
\end{table}

\subsection{Feature Extraction and Cost Volumes}
\label{sec:feature}

ESMStereo feature extraction consists of an encoder and a decoder part. The encoder part uses an Efficient B2 \cite{tan2021efficientnetv2} backbone for ESMStereo-L and ESMStereo-M and MobileNet V2 \cite{26_mobilestereonet} for ESMStereo-S to rapidly generate down-scaled features and the decoder part is comprised of transpose convolutional layers to up-sample the encoded features to \(\frac{1}{4}\), \(\frac{1}{8}\), and \(\frac{1}{16}\) of the original resolution. The features extracted from the final feature extraction layer are used to construct two alternative cost-volume representations: (i) a group-wise correlation cost volume and (ii) a norm-correlation cost volume, defined by Eq.\ref{eq22} and Eq.\ref{eq3}, respectively. These cost-volume formulations are evaluated as separate model configurations throughout this work and are not used simultaneously within a single network instance. For a given ESMStereo variant, only one of the two cost-volume representations is constructed and passed to the subsequent aggregation network.The resulting cost-volume dimensions are \([1, D_{max}/4, H/4, W/4]\), \([1, D_{max}/8, H/8, W/8]\), and \([1, D_{max}/16, H/16, W/16]\) for ESMStereo-L, ESMStereo-M, and ESMStereo-S, respectively, where \(D_{max}\) denotes the maximum disparity.
\setlength{\parindent}{0pt}
\vspace{5mm}
\par
For the group-wise correlation cost volume, the feature map is split to \(N_g\) groups along the channel dimension. Considering the number of feature map channels as \(N_c\), \(g\)th feature group as \(f_l^g\) and \(f_r^g\), the group-wise correlation cost volume can be formed using the dot product as Eq.\ref{eq22} \cite{GWC}. For the norm correlation volume, \(\langle , \rangle\) denotes the inner product, \(d\) is the disparity index, \((x, y)\) represents the pixel coordinate \cite{11_cgistereo}. 

\begin{equation} 
\label{eq22}
\begin{split}
C_{gw-corr}(d, x, y, g) = \frac{N_g}{N_c} \langle {f_l^g(x,y), f_r^g(x-d,y)} \rangle
\end{split}
\end{equation}
{\setlength{\parindent}{0cm}

\begin{equation} 
\label{eq3}
\begin{split}
C_{norm-corr}(:, d, x, y) = \frac{\langle {f_l(:,x,y), f_r(:,x-d,y)}\rangle}{||f_l(:,x,y)||_2 . ||f_r(:,x-d,y)||_2} 
\end{split}
\end{equation}

For aggregating the constructed cost volume, a 3D hourglass network, with the architecture detailed in Table \ref{T2:agg}, is employed. The output is the estimation of the initial disparity map, where \(i=8\) for all ESMStereo versions and \(j=4, 8, 16\) for ESMStereo-S, ESMStereo-M, and ESMStereo-L. Compared to other methods, ESMStereo utilizes significantly fewer channels in the aggregation block which results in reducing the overall computational burden.

\begin{table}[]
\caption{3D aggregation block architecture}
\begin{adjustbox}{width=1\textwidth}
\centering
\begin{tabular}{l l l}

\hline

\multirow{1}{*}{ID} & \multirow{1}{*}{Layer} &  \multirow{1}{*}{Output Size} \\\hline \hline

1 & Input: Cost Volume & \(B \times 1 \times D_{max}/j \times H/j \times W/j\)  \\ \hline 
2 & From (1) Conv3\(\times\)3, BN, GELU & \(B \times j \times D_{max}/j \times H/j \times W/j\)\\
3 & Conv3\(\times\)3, BN, GELU & \(B \times (i+j) \times D_{max}/(2*i) \times H/(2*i) \times W/(2*i)\)\\
4 & Conv3\(\times\)3, BN, GELU & \(B \times (i+j) \times D_{max}/(2*i) \times H/(2*i) \times W/(2*i)\)\\

5 & Conv3\(\times\)3, BN, GELU & \(B \times (i+2*j) \times D_{max}/(4*i) \times H/(4*i) \times W/(4*i)\)\\
6 & Conv3\(\times\)3, BN, GELU & \(B \times (i+2*j) \times D_{max}/(4*i) \times H/(4*i) \times W/(4*i)\)\\

7 & Conv3\(\times\)3, BN, GELU & \(B \times (i+4*j) \times D_{max}/(8*i) \times H/(8*i) \times W/(8*i)\)\\
8 & Conv3\(\times\)3, BN, GELU & \(B \times (i+4*j) \times D_{max}/(8*i) \times H/(8*i) \times W/(8*i)\)\\

9 & Deconv3\(\times\)3, BN, GELU & \(B \times (i+2*j) \times D_{max}/(4*i) \times H/(4*i) \times W/(4*i)\)\\
10 & From (6) and From (9): Concatenation & \(B \times (2*i+4*j) \times D_{max}/(4*i) \times H/(4*i) \times W/(4*i)\)\\
11 & Conv3\(\times\)3, BN, GELU & \(B \times (i+2*j) \times D_{max}/(4*i) \times H/(4*i) \times W/(4*i)\)\\

12 & Deconv3\(\times\)3, BN, GELU & \(B \times (i+j) \times D_{max}/(2*i) \times H/(2*i) \times W/(2*i)\)\\
13 & From (4) and From (12): Concatenation & \(B \times (2*i+2*j) \times D_{max}/(2*i) \times H/(2*i) \times W/(2*i)\)\\

14 & conv3\(\times\)3, BN, GELU & \(B \times (i+j) \times D_{max}/(2*i) \times H/(2*i) \times W/(2*i)\)\\

15 & Deconv3\(\times\)3 & \(B \times 1 \times D_{max}/i \times H/i \times W/i\)\\ \hline

\end{tabular}
\end{adjustbox}
\label{T2:agg}
\end{table}

\subsection{Disparity Regression}
\label{sec:reg}
The output of the aggregation network is regularized by selecting the top-$k$ values at every pixel as described in \cite{12_coex} with \(k=1\) to produce an initial disparity map at \(d_{\frac{1}{8}}\) and \(d_{\frac{1}{16}}\), and \(k=2\) at \(d_{\frac{1}{4}}\). To produce the full resolution disparity map \(d_0\) is progressively up-sampled using ESM modules in multiple stages.

\subsection{Loss Function}
\label{sec:loss}

ESMStereo models are trained end-to-end and supervised by the weighted loss function described in Eq.\ref{eq4}. For example, the ESMStereo-8 model has three stacked ESM modules \(N=3\), it generates three disparity maps at \(d_i = \{ d_{\frac{1}{4}}, d_{\frac{1}{2}}, d_{\frac{1}{1}}\}\) resolution, where \(i\) denotes the disparity prediction index produced at each progressive upsampling stage. Therefore, we supervise these multi-scale disparity maps by interpolating the ground truth to match the dimensions, \textit{i.e.}, \(d_i^{gt} = \{d_{\frac{1}{4}}^{gt}, d_{\frac{1}{2}}^{gt}, d_{\frac{1}{1}}^{gt}\}\). In  Eq.\ref{eq4}, \(\lambda_i = \{1, \frac{1}{6}, \frac{1}{10}\}\) are the reduction factors and \(smooth_{L_1}\) expressed as Eq.\ref{smooth}.

\begin{equation} 
\label{eq4}
\begin{split}
L = \sum_{i=0}^N\lambda_i smooth_{L_1}(d_i-d_i^{gt})
\end{split}
\end{equation}

\begin{equation} 
\label{smooth}
\begin{split}
smooth_{L_1}(x) = \begin{cases} \frac{x^2}{2} & \text{, $|x| < 1$} \\ |x|-0.5 &\text{, $|x| \geq 1$}
\end {cases} 
\end{split}
\end{equation}

\section{Experiment}
In this section, four datasets are utilized to evaluate the performance of ESMStereo models and analyze their generalization capability. These datasets are widely used for benchmarking models in the stereo matching domain \cite{xu2023iterative, zhao2023high, 10_hitnet}.

\subsection{Datasets and Evaluation Metrics}
\label{sec:data}
\textbf{SceneFlow} \cite{32_dispnet} is a synthetic dataset comprising 35,454 training image pairs and 4,370 testing image pairs, each with a resolution of 960×540. Performance on SceneFlow is evaluated using the End-Point Error (EPE), as defined in Eq.\ref{eq5}, where \((x, y)\) represents the pixel coordinates, \(d\)  denotes the estimated disparity, \(d_{gt}\) is the ground truth disparity and \(N\) is the effective pixel number in one disparity image. Another evaluation metric for SceneFlow is the disparity outlier (D1), which refers to pixels with errors exceeding max \(max(3px, 0.05d_{gt})\).  Due to its large size, SceneFlow is commonly used for pre-training stereo matching networks before fine-tuning on real-world benchmarks.

\begin{equation} 
\label{eq5}
\begin{split}
EPE = \frac{\sum_{(x, y)}|d(x, y) - d_{gt}(x, y)|}{N}
\end{split}
\end{equation}

{\setlength{\parindent}{0cm}

\textbf{KITTI} includes two benchmarks: KITTI 2012 \cite{kitti_12} and KITTI 2015 \cite{kitti_15}. KITTI 2012 comprises 194 training stereo image pairs and 195 testing image pairs, while KITTI 2015 includes 200 training stereo image pairs and 200 testing image pairs. These datasets feature real-world driving scenes and provide sparse ground-truth disparity measured by LiDAR.
{\setlength{\parindent}{0cm}
\vspace{5mm}
\par
\textbf{ETH3D} \cite{ETH3D} consists of 27 training and 20 testing grayscale image pairs, with sparse ground-truth disparity. The disparity range in ETH3D extends from 0 to 64. Performance evaluation on the ETH3D dataset is based on the percentage of pixels with errors larger than 1 pixel (referred to as ``bad 1.0").
{\setlength{\parindent}{0cm}
\vspace{5mm}
\par
\textbf{Middlebury 2014} \cite{midburry} is a dataset comprising 15 training and testing indoor image pairs, available at full, half, and quarter resolutions. Evaluation on this dataset is based on the percentage of pixels with errors larger than 2 pixels (referred to as ``bad 2.0"). Additionally, bad-\(\sigma\) error can be defined as Eq.\ref{eq6}.

\begin{equation} 
\label{eq6}
\begin{split}
bad-\sigma = \frac{\sum_{(x, y)}|d(x, y) - d_{gt}(x, y)| > \sigma}{N} * 100 \%
\end{split}
\end{equation}

\subsection{Training Details}

ESMStereo is implemented in PyTorch and evaluated on NVIDIA RTX 4070 SUPER (12GB), and Jetson AGX Orin (64GB). The optimization is performed using the ADAMW algorithm \cite{loshchilov2017decoupled} with \(\beta_1 = 0.9\) and \(\beta_2 = 0.999\), batch size of 4, maximum disparity of 192 and \(256 \times 512\) resolution (crop size). The data are augmented through brightness, contrast, and saturation adjustments. Furthermore, the feature extraction backbone is trained from scratch for all ESMStereo variants, including both the encoder and decoder components. The encoder employs EfficientNet-B2 \cite{tan2021efficientnetv2} for ESMStereo-L and ESMStereo-M, and MobileNetV2 \cite{26_mobilestereonet} for ESMStereo-S to generate compact multi-scale representations, while the decoder uses transposed convolution layers to progressively upsample features to \(\frac{1}{4}\), \(\frac{1}{8}\), and \(\frac{1}{16}\) of the original image resolution. The complete training is conducted in two stages on the NVIDIA RTX 4070 SUPER: initial training on the synthetic SceneFlow dataset for 60 epochs, followed by fine-tuning for an additional 80 epochs. During the initial training phase, the learning rate is set to 0.001 and decayed by a factor of 2 after epochs 20, 32, 40, 48, and 56. In the fine-tuning phase, the learning rate starts at 0.0002 and is decayed by a factor of 2 after epochs 20, 30, 40, 50, 60, and 70. 
\setlength{\parindent}{0pt}
\vspace{5mm}
\par
For training on the KITTI datasets, the model pre-trained on SceneFlow is fine-tuned for 600 epochs using the combined KITTI 2012 and KITTI 2015 training sets. During this phase, the learning rate is initially set to 0.001 and reduced to 0.0001 at the 300th epoch. Additionally, zero-shot generalization results on KITTI, ETH3D, and Middlebury are obtained using the model trained exclusively on SceneFlow.

\subsection{Ablation Study}
\label{abl}

To evaluate the effectiveness of different cost volumes, we conducted an ablation experiment on the SceneFlow dataset. The models were trained using group-wise correlation (denoted as -gwc) and norm-correlation (denoted as -nc) cost volumes at \(\frac{1}{4}\), \(\frac{1}{8}\) and \(\frac{1}{16}\) resolutions, corresponding to the large (L), medium (M), and small (S) model versions, respectively. As shown in Table \ref{T22}, the results demonstrate that group-wise correlation consistently improves accuracy across all model scales, indicating that preserving localized channel-wise matching relationships is more effective than global norm-based similarity for compact cost volumes. The best accuracy is obtained by ESMStereo-L-gwc, which achieves an EPE of 0.53 pixels at an inference time of 23 ms. The fastest model, ESMStereo-S-gwc, delivers an EPE of 1.10 pixels with an inference time of 8.6 ms.

\setlength{\parindent}{0pt}
\vspace{5mm}
\par

Considering sensitivity analysis for models using group-wise correlation cost volume, reducing the cost volume resolution from 1/4 to 1/16 decreases runtime by approximately 3x while increasing EPE from 0.53 to 1.10 pixels which illustrates the tradeoff between geometric detail preservation and computational efficiency. 

\setlength{\parindent}{0pt}
\vspace{5mm}
\par

To investigate lightweight feature extraction backbones for real-time stereo matching, MobileNetV2 \cite{sandler2018mobilenetv2} is employed in the small ESMStereo variants to reduce computational complexity and improve inference speed. The resulting ESMStereo-S-gwc and ESMStereo-S-nc models achieve inference rates exceeding 100 FPS that indicates their suitability for deployment on resource-constrained edge devices while maintaining competitive disparity estimation accuracy.

\begin{table}[H]
\centering
\caption{Ablation study on SceneFlow dataset. The bold font indicates the best EPE.}

\begin{adjustbox}{width=1\textwidth}
\small
\begin{tabular}{C{1.8cm} | L{2.8cm}  C{2cm}  C{1.7cm}  C{1.8cm} L{2.5cm} | C{1.2cm} C{1cm}  C{1.6cm} C{1.6cm} C{1.6cm}}
\hline

\multirow{2}{*}{Architecture} & \multirow{2}{*}{\shortstack[c]{Model \\ Name}} & \multirow{2}{*}{\shortstack[c]{Cost Volume \\ Resolution}} & \multirow{2}{*}{\shortstack[c]{Norm \\ Correlation}} & \multirow{2}{*}{\shortstack[c]{Group-Wise \\ Correlation}} & \multirow{2}{*}{\shortstack[c]{Backbone}} &
\multirow{2}{*}{EPE[px]} &  \multirow{2}{*}{D1[\%]} & \multirow{2}{*}{\shortstack[c]{Time[ms] \\ RTX 4070}} & \multirow{2}{*}{\shortstack[c]{FLOPs[G]}}\\
& & & & & & &&& \\\hline \hline

\parbox[t]{2mm}{\multirow{6}{*}{\rotatebox[origin=c]{90}{ESMStereo}}}
& ESMStereo-S-nc & $1/16$ &  \Checkmark  &  & MobileNet V2 & 1.21 &  5.0  & 8.8 & 9.1\\ 
& ESMStereo-S-gwc & $1/16$ &   &  \Checkmark &  MobileNet V2 & 1.10 &  4.6 & 8.6 & 9.4\\ 
& ESMStereo-M-nc & $1/8$&  \Checkmark   &  &  EfficientNet B2  & 0.80 &  3.1 & 14 & 30\\
& ESMStereo-M-gwc & $1/8$&     &\Checkmark  &  EfficientNet B2  & 0.77 & 3.0 & 14 & 31\\ 
& ESMStereo-L-nc & $1/4$ &   \Checkmark  &  & EfficientNet B2 & 0.55 & 2.0  & 24 & 58\\ 
& ESMStereo-L-gwc & $1/4$ &  & \Checkmark  &  EfficientNet B2 & \textbf{0.53} & \textbf{1.9}  & 26 & 69\\\cline{2-10}
\hline

\end{tabular}	
\end{adjustbox}
\label{T22}
\end{table}

Table \ref{T22-al} reflects the complementary roles of the proposed components from a stereo matching perspective. The FMBlock introduces structured feature mixing through shuffle-based aggregation, which enhances the discriminative power of matching features while maintaining computational efficiency. This improves the quality of the initial cost representation to achieve reliable disparity hypotheses before refinement.
\setlength{\parindent}{0pt}
\vspace{5mm}
\par
The most significant improvement, however, is obtained by the feature-guided hourglass refinement module. In stereo matching, initial cost volumes often contain ambiguity in textureless or repetitive regions, and accurate disparity estimation requires progressive context aggregation and spatial consistency enforcement. The refinement module addresses this by integrating image-guided contextual features with disparity features across multiple scales that enables the network to progressively resolve local ambiguities and recover fine geometric structures. As a result, the combination of structured feature enhancement (FMBlock) and iterative image-guided refinement yields a substantial reduction in EPE, as shown in Table \ref{T22-al}.

\setlength{\parindent}{0pt}
\vspace{5mm}
\par

We note that image-feature fusion is inherently embedded within the refinement process and cannot be cleanly isolated as an independent ablation component without significantly altering the network topology and intermediate feature dimensionalities. For this reason, and to ensure a consistent and meaningful comparison, the fusion mechanism is evaluated jointly with the refinement module in the ablation study.

\begin{table}[H]
\centering
\caption{Ablation study on the contribution of proposed components on SceneFlow dataset. The bold font indicates the best EPE.}

\begin{adjustbox}{width=1\textwidth}
\small
\begin{tabular}{L{1.8cm} | C{2.5cm}  C{1.7cm}  C{3.5cm} | C{1.2cm}   C{1.6cm} }
\hline

\multirow{2}{*}{Component} & \multirow{2}{*}{\shortstack[c]{Disparity feature \\ extraction}} & \multirow{2}{*}{\shortstack[c]{FMBlock}} & \multirow{2}{*}{\shortstack[c]{Feature-guided \\ hourglass Refinement}} &
\multirow{2}{*}{EPE[px]}  & \multirow{2}{*}{\shortstack[c]{Time[ms] \\RTX 4070}} \\
& & & & &  \\\hline \hline

\parbox[t]{2mm}{\multirow{4}{*}{\shortstack[c]{ESMStereo-\\L-gwc}}}
 & \Checkmark&    &  & 1.20  & \textbf{23.5}\\ 
& \Checkmark &  \Checkmark  &   & 0.92  & 24.1\\ 
&  \Checkmark &  \Checkmark   &  \Checkmark &  \textbf{0.53}  & 26\\
\cline{2-6}
\hline

\end{tabular}	
\end{adjustbox}
\label{T22-al}
\end{table}

The runtime breakdown in Table \ref{runtime_normalized} illustrates the computational contribution of each component in ESMStereo. As shown, the feature extraction stage accounts for the majority of the computational cost across all model variants, indicating that it is the dominant factor in overall runtime. In contrast, the cost volume construction and aggregation modules remain relatively lightweight, particularly in the smaller variants. A key observation is that the proposed ESM module, while introducing an additional refinement stage, enables the use of significantly more compact cost volume and aggregation designs without sacrificing accuracy. This demonstrates that ESM effectively shifts the computational burden from expensive 3D processing to a more efficient 2D refinement paradigm, which results in a balanced and efficient architecture.

\begin{table}[H]
\centering
\caption{Runtime breakdown of ESMStereo components on RTX 4070 Super.}
\label{runtime_normalized}

\begin{adjustbox}{width=0.95\textwidth}
\small
\begin{tabular}{l|c|c|c|c|c}
\hline
Architecture & Feature Extraction & Cost Volume & Aggregation & ESM Modules & Total [ms] \\
\hline
ESMStereo-S-gwc & 6.85 & 0.08 & 0.10 & 1.57 & \textbf{8.60} \\
ESMStereo-M-gwc & 10.86 & 0.62 & 0.25 & 2.27 & \textbf{14.00} \\
ESMStereo-L-gwc & 11.80 & 6.22 & 3.18 & 4.81 & \textbf{26.01} \\
\hline
\end{tabular}
\end{adjustbox}

\end{table}

\subsection{Comparisons with State-of-the-art}

In this section, the proposed architectures are compared with state-of-the-art real-time stereo matching methods. Since prior works report inference times on different GPU platforms (e.g., RTX 3090, Titan X, Tesla V100, and RTX 2080Ti), all runtime results are presented together with the corresponding hardware specifications to ensure transparency and fair interpretation. Accuracy metrics such as EPE and D1 are inherently hardware-independent, while inference times may vary depending on GPU architecture, memory bandwidth, and tensor core utilization. In particular, GPUs such as the RTX 3090 and RTX 4070 Super differ in architectural generation and compute characteristics, which can lead to non-trivial variations in runtime even for identical models. Therefore, direct cross-hardware comparisons should be interpreted with caution, especially when evaluating real-time performance claims. To complement runtime measurements, we also report computational complexity in terms of FLOPs, which provides a hardware-agnostic indicator of algorithmic efficiency by reflecting the underlying computational cost independent of specific GPU optimizations, memory hierarchies, or kernel implementations. Accordingly, we consider both FLOPs and reported runtimes jointly to assess real-time capability in a more balanced and reliable manner.

\setlength{\parindent}{0pt}
\vspace{5mm}
\par

\textbf{SceneFlow}. ESMStereo is designed with two different cost volume resolutions to balance accuracy and speed while ensuring real-time performance. Table \ref{T2} demonstrates the performance of ESMStereo-{S, M, L}-gwc on the SceneFlow test set, compared to other state-of-the-art real-time approaches. The results show that ESMStereo-L-gwc achieves \(EPE = 0.53 px\) on the SceneFlow test set, outperforming approaches such as LightStereo-L \cite{guo2024lightstereo}, RT-IGEV++ \cite{xu2024igev++} and Fast-ACVNet+ \cite{25_acvnet}. Considering super lightweight stereo matching networks such as SADSNet \cite{wu2023towards} with the speed of above 100 FPS, ESMStereo-S-gwc delivers better accuracy with the EPE of 1.10 pixel at 116 FPS.

\setlength{\parindent}{0pt}
\vspace{5mm}
\par

Considering different upsampling strategies, ESMStereo-L-gwc achieves 0.53 px EPE on SceneFlow, outperforming RTSMNet-c8 \cite{35_rtsmnet} (0.71 px) and FBPGNet \cite{WEN2022116636} (1.19 px), which both employ progressive upsampling designs. This corresponds to approximately 25.4\% and 55.5\% relative improvements in accuracy, respectively, while maintaining competitive runtime across different hardware settings. On the other hand, ESMStereo-L-gwc also consistently outperforms BGNet \cite{9_bgnet}, which utilizes bilateral grid-based upsampling, Fast-ACVNet \cite{25_acvnet} with bilinear interpolation-based upsampling, and CGIStereo \cite{11_cgistereo} which relies on deconvolution-based refinement. These results further demonstrate that the proposed ESM module provides a more effective and generalizable refinement mechanism compared to a range of representative upsampling strategies in stereo matching.
\setlength{\parindent}{0pt}
\vspace{5mm}
\par
Comparison of GFLOPs in Table \ref{T2} shows that ESMStereo achieves a strong accuracy–efficiency trade-off. In particular, ESMStereo-L-gwc obtains the lowest EPE (0.53 px) while maintaining only 69 GFLOPs and 6.8M parameters which demonstrates a compact and computationally efficient design. Compared with methods such as IINet \cite{li2024iinet} (90 GFLOPs, 20M params) and Fast-ACVNet+ \cite{25_acvnet} (93 GFLOPs), ESMStereo-L-gwc achieves better accuracy with lower computational cost. In contrast, heavier models like FADNet++ \cite{wang2021fadnet++} (148 GFLOPs, 124M params) require significantly more computation but yield inferior accuracy. These results indicate that the proposed method improves efficiency in a hardware-independent manner, as reflected by FLOPs and parameter counts rather than GPU-specific runtime.

\begin{table}[H]
\centering
\caption{Evaluation on SceneFlow Dataset.}
\begin{tabular}{l|c|c|c|c|c}
\hline

Method (Real-Time) & EPE [px] & Time [ms] & FLOPs [G] & Params [M] & GPU \\
\hline \hline

SADSNet-M-N7 \cite{wu2023towards} & 1.16 & 8.5 & 2.6 & 0.4 & RTX 3090 \\
SADSNet-L-N7 \cite{wu2023towards} & 0.90 & 13 & 9.6 & 1.5 & RTX 3090 \\
StereoNet \cite{1_stereonet} & 1.10 & 15 & 85 & 0.4 & Titan X \\
LightStereo-S \cite{guo2024lightstereo} & 0.73 & 17 & 22 & 3.4 & RTX 3090 \\
ADCPNet \cite{8_adcpnet} & 1.48 & 20 & -- & -- & 2080Ti \\
Fast-ACVNet \cite{25_acvnet} & 0.64 & 22 & 79 & 3.02 & RTX 3090 \\
BGNet \cite{9_bgnet} & 1.17 & 25 & 45 & 3.8 & RTX 3090 \\
IINet \cite{li2024iinet} & 0.54 & 26 & 90 & 20 & RTX 3090 \\
Coex \cite{12_coex} & 0.68 & 27 & 53 & 2.27 & RTX 3090 \\
Fast-ACVNet+ \cite{25_acvnet} & 0.59 & 27 & 93 & 3.2 & RTX 3090 \\
RTSMNet-c8 \cite{35_rtsmnet} & 0.71 & 28 & - & 1.1 & 2080Ti \\
CGIStereo \cite{11_cgistereo} & 0.64 & 29 & 58 & 2.9 & RTX 3090 \\
EBStereo \cite{20_ebstereo} & 0.63 & 29 & -- & -- & RTX 3090 \\
FADNet++ \cite{wang2021fadnet++} & 0.76 & 33 & 148 & 124 & Tesla V100 \\
LightStereo-L \cite{guo2024lightstereo} & 0.59 & 37 & 36 & 7.6 & RTX 3090 \\
RT-IGEV++ \cite{xu2024igev++} & 0.55 & 42 & -- & -- & RTX 3090 \\
FBGNet-4 \cite{WEN2022116636} & 1.19 & 44 & -- & 1.3 & RTX 3090 \\
\hline

\textbf{ESMStereo-S-gwc} & 1.10 & 8.6 & 9.4 & 1.8 & RTX 4070S \\
\textbf{ESMStereo-M-gwc} & 0.77 & 14 & 31 & 6.3 & RTX 4070S \\
\textbf{ESMStereo-L-gwc} & \textbf{0.53} & 26 & 69 & 6.8 & RTX 4070S \\
\hline

\end{tabular}
\label{T2}
\end{table}

{\setlength{\parindent}{0cm}
\vspace{5mm}
\par
\textbf{KITTI 2012 and 2015}. Table  \ref{T3} presents the official results for ESMStereo models on the KITTI 2012 and KITTI 2015 datasets. Among the real-time methods evaluated on KITTI 2012, ESMStereo-L-gwc achieves the best results. On KITTI 2015, ESMStereo-L-gwc exhibits the best performance for the percentage of outliers averaged over background regions (D1-bg) and ranks second for the percentage of stereo disparity outliers in the reference frame (D1-all). Furthermore, although ESMStereo-S-gwc and ESMStereo-M-gwc achieve lower accuracy than some state-of-the-art methods with inference times exceeding 20 ms, they remain highly competitive among methods operating within similar runtime ranges. For instance, ESMStereo-S-gwc achieves competitive accuracy with an inference time of 8.6 ms compared to HRSNet \cite{huang2023real} at 7.0 ms, ADCPNet \cite{8_adcpnet} at 7.0 ms, and SADSNet-M-N7 \cite{wu2023towards} at 8.3 ms. Similarly, ESMStereo-M-gwc achieves a D1-all error of 2.34\% on KITTI 2015 with a 14 ms inference time, outperforming SADSNet-L-N7 \cite{wu2023towards} and P3SNet \cite{33_p3snet}, which report D1-all errors of 2.74\% and 5.05\% with inference times of approximately 12 ms, respectively.

\subsection{Performance on Edge Computing Devices}

The ESMStereo models are deployed on the NVIDIA Jetson AGX Orin platform with 64 GB of memory. To optimize inference performance on edge hardware, the models are converted into TensorRT engines using FP16 (16-bit floating-point) precision. Inference performance is evaluated in terms of frames per second (FPS) using the trtexec tool over 200 consecutive runs. We additionally evaluated the impact of FP16 conversion on accuracy using the SceneFlow benchmark. The results show no observable degradation in EPE compared to the original implementation. This is mainly due to two factors: (i) ESMStereo relies primarily on convolutional and correlation-based operations that are well-supported under FP16 precision on modern Tensor Cores, and (ii) the model does not involve aggressive quantization or nonlinear operations that are sensitive to reduced numerical precision. As shown in Table \ref{agx}, the ESMStereo-S-gwc and ESMStereo-M-gwc models achieve inference rates exceeding 90 FPS and 25 FPS, respectively, while maintaining identical benchmark accuracy to the original FP32 baseline.

\begin{table}[H]
\centering
\caption{Comparison of performance on AGX Orin 64GB with RTX 4070. Results are reported using the model trained only on SceneFlow dataset and evaluated on KITTI 2015 training set}

\begin{adjustbox}{width=0.5\textwidth}
\small
\begin{tabular}{l|c c| c c}
\hline
\multirow{2}{*}{Architecture} & \multicolumn{2}{c|}{AGX Orin 64GB}  &   \multicolumn{2}{c}{RTX 4070S} \\

& \multirow{1}{*}{D1[\%]} & \multirow{1}{*}{FPS[Hz]} & \multirow{1}{*}{D1[\%]} & \multirow{1}{*}{FPS[Hz]} \\ \hline
ESMStereo-S-gwc &  10.8   & 91 &  10.8 & 116 \\ 
ESMStereo-M-gwc &  8.2   & 29 &  8.2 & 71 \\
ESMStereo-L-gwc &  5.5   & 8.4 &  5.5 & 38 \\\hline

\end{tabular}	
\end{adjustbox}
\label{agx}
\end{table}

\begin{table}[H]
\caption{Evaluation on KITTI Datasets. The methods are categorized based on whether their design focus is primarily for accuracy or speed. The bold font indicates best performance and underlined values indicate the second best. Our results are reported using RTX 4070S.}
\centering
\begin{adjustbox}{width=1\textwidth}
\small

\begin{tabular}{c|l| c c c c c c| c c c| c}

\multicolumn{2}{c|}{} & \multicolumn{6}{c}{KITTI 2012}  &  \multicolumn{3}{|c|}{KITTI 2015} &  \\\hline \hline

\multirow{1}{*}{Target} & \multirow{1}{*}{Method} &  \multirow{1}{*}{3-Noc} & \multirow{1}{*}{3-All} & \multirow{1}{*}{4-Noc} & \multirow{1}{*}{4-all} & \multirow{1}{*}{EPE noc} & \multirow{1}{*}{EPE all} & \multirow{1}{*}{D1-bg} & \multirow{1}{*}{ D1-fg} & \multirow{1}{*}{D1-all} & \multirow{1}{*}{Time[ms]} \\\hline 

\parbox[t]{2mm}{\multirow{10}{*}{\rotatebox[origin=c]{90}{Accuracy}}} 
 & CFNet \cite{shen2021cfnet} &1.23 & 1.58&0.92 & 1.18&0.4 &0.5 &1.54 &3.56 &1.88 & 180\\  
 & IGEV-Stereo \cite{xu2023iterative} &1.12 & 1.44& 0.88 & 1.12&0.4 &0.4 &1.38 &2.67 &1.59 & 180\\  
 & ACVNet \cite{25_acvnet}&1.13 &1.47 & 0.86&1.12 &0.4 &0.5 &1.37 & 3.07&1.65&200  \\ 
 & LEAStereo \cite{cheng2020hierarchical} &1.13 &1.45 &0.83 &1.08 &0.5 &0.5 &1.40 &2.91 &1.65&300\\   
 & EdgeStereo-V2 \cite{song2020edgestereo}&1.46 &1.83&1.07&1.34 & 0.4&0.5 & 1.84&3.30 & 2.08&320 \\  
 & CREStereo \cite{18_deeppruner} &1.14 &1.46 &0.90 & 1.14& 0.4&0.5 &1.45 &2.86 &1.69 &410 \\  
 & SegStereo \cite{yang2018segstereo} &1.68 &2.03 &1.25 &1.52 &0.5 &0.6 & 1.88&4.07 &2.25 &600 \\  
 & SSPCVNet \cite{wu2019semantic} &1.47 &1.90 &1.08 &1.41 &0.5 &0.6 &1.75 &3.89 &2.11 &900 \\  
 & CSPN \cite{cheng2019learning} &1.19 &1.53 &0.93 & 1.19&- &- &1.51 &2.88 &1.74 &1000 \\  
 & GANet \cite{22_ganet} & 1.19&1.60 & 0.91&1.23 & 0.4&0.5 &1.48 &3.46&1.81 &1800 \\  
 & LaC+GANet \cite{liu2022local} &1.05&1.42 &0.80 &1.09 &0.4 &0.5&1.44 &2.83 &1.67 &1800 \\\hline
 
 \hline 
\parbox[t]{2mm}{\multirow{18}{*}{\rotatebox[origin=c]{90}{Speed (Real-Time)}}} 
&GWDRNet-S \cite{huang2023real}  &2.26 &3.17 &- &- &- & -&- &- &3.64 &\textbf{6.0}\\ 
&HRSNet \cite{huang2023real}  &- &- &- &- &- & -&3.27 &7.58 &3.98 &7.0\\ 
&ADCPNet \cite{8_adcpnet}  &3.14 &3.84 &- &- &- & -&- &- &3.98 &7.0\\ 
&SADSNet-M-N7  \cite{wu2023towards} &2.39 &2.98 &- &- &- & -&- &- &3.11 &8.3\\ 
&SADSNet-L-N7  \cite{wu2023towards} &2.12 &2.66 &- &- & - & -&- &- &2.74 &12\\ 
&P3SNet \cite{33_p3snet} &3.65 &4.46 &2.51 &3.18 & - & -&4.4 &8.28 &5.05 &12\\ 
& LightStereo-S \cite{guo2024lightstereo} &1.88 &2.34 &1.30 &1.65 &0.6 & 0.6&2.00 &3.80 &2.30 &17\\  
& RTSMNet \cite{35_rtsmnet} &- &- &- &- &- & -&3.44 &6.08 &3.88 &19\\  
& HITNet \cite{10_hitnet} & 1.41&1.89 &1.14 &1.53 &0.4 &0.5 &1.74& \underline{3.20}&1.98 &20\\ 
& CGI-Stereo \cite{11_cgistereo} &1.41 &1.76 &1.05 &1.30 &0.5 & 0.5&1.66 &3.38 &1.94 &29\\  
& CoEx \cite{12_coex} & 1.55&1.93 &1.15 &1.42 & 0.5&0.5 &1.79 &3.82 &2.13 &33\\  
& LightStereo-L \cite{guo2024lightstereo} &1.55 &1.87 &1.10 &1.33 &0.5 & 0.5&1.78 &\textbf{2.64} &1.93 &34\\ 
& BGNet+ \cite{9_bgnet} &1.62 & 2.03& 1.16&1.48 &0.5 &0.6 &1.81 &4.09 &2.19 &35\\  
& Fast-ACVNet+ \cite{xu2022attention} & 1.45&1.85 &1.06 & 1.36&0.5 & 0.5& 1.70&3.53 &2.01 &45\\  
& RT-IGEV++ \cite{xu2024igev++} & \underline{1.29} &1.68 &- & -&0.4 & 0.5& \underline{1.48}&3.37 &\textbf{1.79} &48\\ 
& DecNet \cite{yao2021decomposition} &- &- & -&- &- &- &2.07 & 3.87& 2.37&50\\
 & MDCNet \cite{chen2021multi} &1.54 &1.97 &- &- &- &- &1.76  &- &2.08 & 50\\  
& DeepPrunerFast \cite{18_deeppruner} &- &- &- &- &- &- &2.32 &3.91 &2.59 &50\\ 
& DispNetC \cite{32_dispnet} &4.11 &4.65 & 2.77& 3.20& 0.9& 1.0&2.21 &6.16 &4.43 &60\\  
& AANet \cite{28_aanet} &1.91 &2.42 &1.46 &1.87 &0.5 &0.6 &1.99 &5.39 &2.55 &62\\  
& DCVSMNet \cite{TAHMASEBI2024129002}  &1.30 &\underline{1.67} &\underline{0.96} &\underline{1.23} &0.5 &0.5 & 1.60 &3.33  &1.89 &  67 \\

& JDCNet \cite{jia2021joint} &1.64 &2.11 &- &- &- &- &1.91  &4.47  &2.33 & 80\\  \hline


& \textbf{ESMStereo-S-gwc}  &3.0 &3.45 &1.98 &2.31 &0.8 &0.8 & 3.06 &4.51  &3.30 &  8.6 \\
& \textbf{ESMStereo-M-gwc}  &1.95 &2.37 &1.36 &1.69 &0.6 &0.6 & 2.09 &3.63  &2.34 &  14 \\
& \textbf{ESMStereo-L-gwc}  &\textbf{1.15} &  \textbf{1.52} &\textbf{0.88} & \textbf{1.16} &0.4 &0.5 & \textbf{1.43} & 3.80 &\underline{1.82}&  26 \\ \hline

\end{tabular}
\end{adjustbox}

\label{T3}
\end{table}
\begin{figure}
\centering
\begin{subfigure}{.22\textwidth}
  \centering
  \captionsetup{font=scriptsize, labelformat=empty} 
  \caption{Left image}
  \includegraphics[width=1\linewidth]{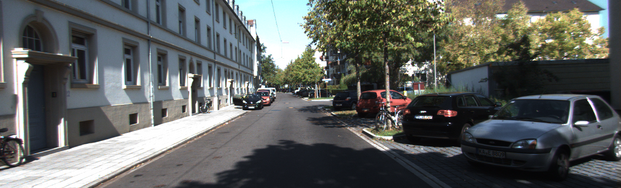}
\end{subfigure}
\begin{subfigure}{.22\textwidth}
  \centering
  \captionsetup{font=scriptsize, labelformat=empty} 
  \caption{ESMStereo-S-gwc}
  \includegraphics[width=1\linewidth]{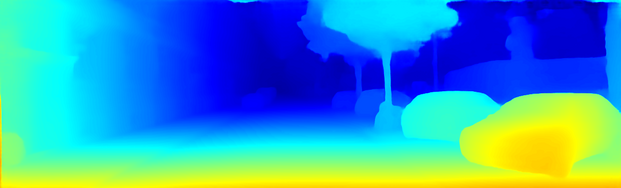}
\end{subfigure}
\begin{subfigure}{.22\textwidth}
  \centering
  \captionsetup{font=scriptsize, labelformat=empty} 
  \caption{ESMStereo-M-gwc}
  \includegraphics[width=1\linewidth]{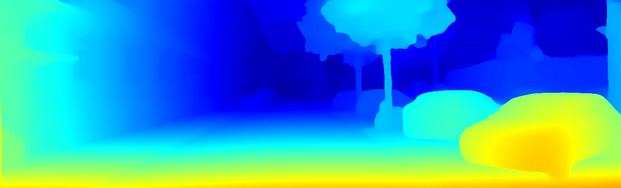}
\end{subfigure}
\begin{subfigure}{.22\textwidth}
  \centering
  \captionsetup{font=scriptsize, labelformat=empty} 
  \caption{ESMStereo-L-gwc}
  \includegraphics[width=1\linewidth]{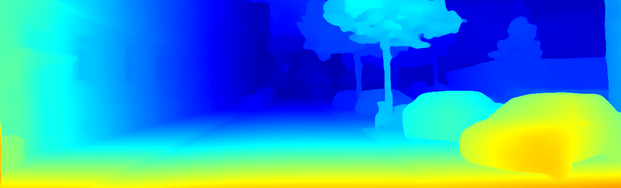}
\end{subfigure}
\hfill
\begin{subfigure}{.22\textwidth}
  \centering
  \includegraphics[width=1\linewidth]{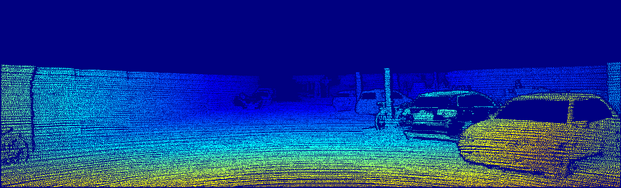}
\end{subfigure}
\begin{subfigure}{.22\textwidth}
  \centering
  \includegraphics[width=1\linewidth]{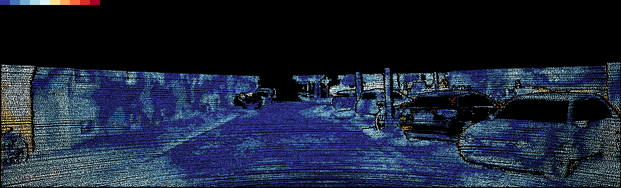}
\end{subfigure}
\begin{subfigure}{.22\textwidth}
  \centering
  \includegraphics[width=1\linewidth]{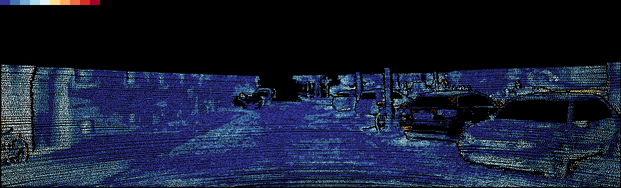}
\end{subfigure}
\begin{subfigure}{.22\textwidth}
  \centering
  \includegraphics[width=1\linewidth]{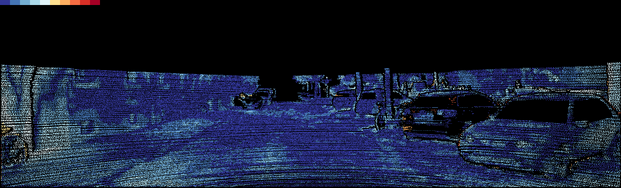}
\end{subfigure}
\begin{subfigure}{.22\textwidth}
  \centering
  \includegraphics[width=1\linewidth]{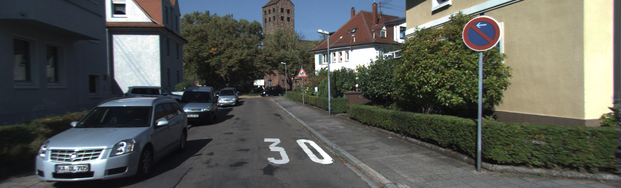}
\end{subfigure}
\begin{subfigure}{.22\textwidth}
  \centering
  \includegraphics[width=1\linewidth]{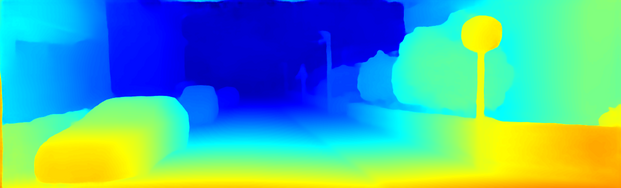}
\end{subfigure}
\begin{subfigure}{.22\textwidth}
  \centering
  \includegraphics[width=1\linewidth]{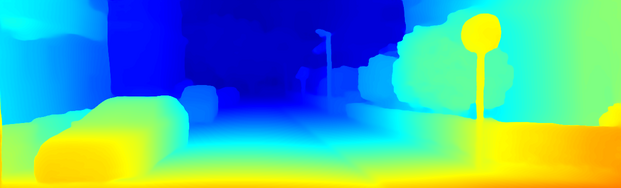}
\end{subfigure}
\begin{subfigure}{.22\textwidth}
  \centering
  \includegraphics[width=1\linewidth]{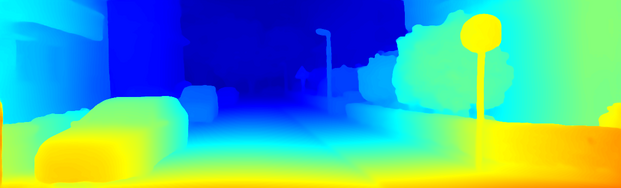}
\end{subfigure}
\hfill
\begin{subfigure}{.22\textwidth}
  \centering
  \includegraphics[width=1\linewidth]{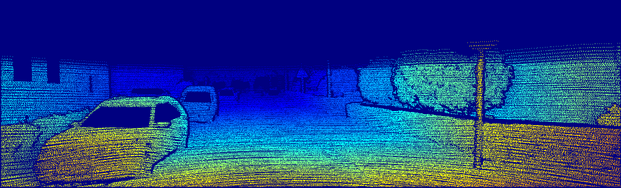}
\end{subfigure}
\begin{subfigure}{.22\textwidth}
  \centering
  \includegraphics[width=1\linewidth]{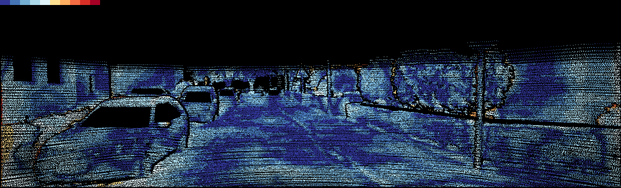}
\end{subfigure}
\begin{subfigure}{.22\textwidth}
  \centering
  \includegraphics[width=1\linewidth]{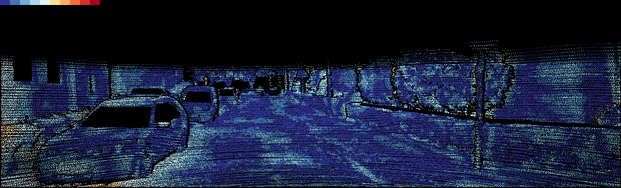}
\end{subfigure}
\begin{subfigure}{.22\textwidth}
  \centering
  \includegraphics[width=1\linewidth]{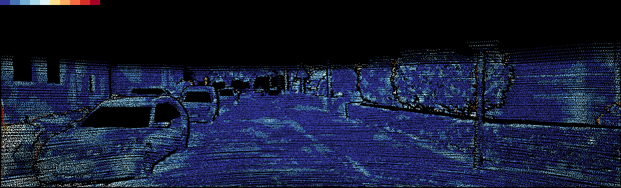}
\end{subfigure}
\begin{subfigure}{.22\textwidth}
  \centering
  \includegraphics[width=1\linewidth]{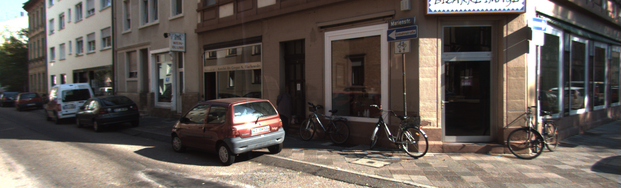}
\end{subfigure}
\begin{subfigure}{.22\textwidth}
  \centering
  \includegraphics[width=1\linewidth]{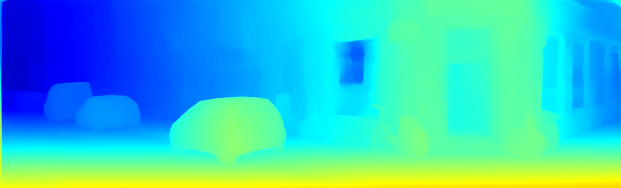}
\end{subfigure}
\begin{subfigure}{.22\textwidth}
  \centering
  \includegraphics[width=1\linewidth]{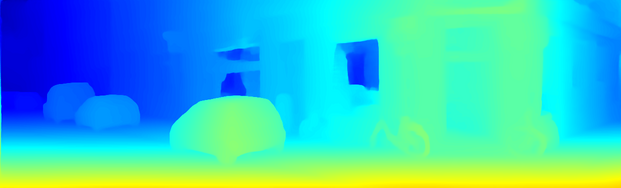}
\end{subfigure}
\begin{subfigure}{.22\textwidth}
  \centering
  \includegraphics[width=1\linewidth]{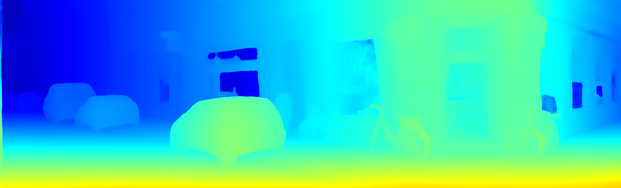}
\end{subfigure}
\hfill
\begin{subfigure}{.22\textwidth}
  \centering
  \includegraphics[width=1\linewidth]{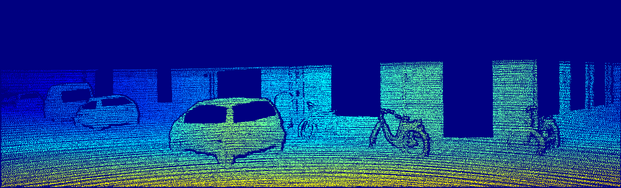}
\end{subfigure}
\begin{subfigure}{.22\textwidth}
  \centering
  \includegraphics[width=1\linewidth]{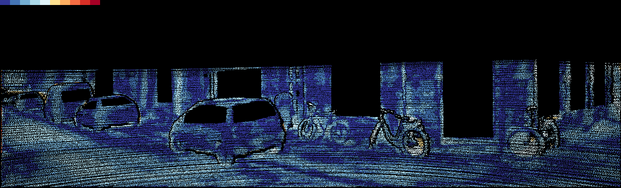}
\end{subfigure}
\begin{subfigure}{.22\textwidth}
  \centering
  \includegraphics[width=1\linewidth]{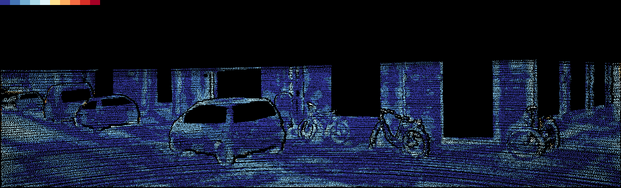}
\end{subfigure}
\begin{subfigure}{.22\textwidth}
  \centering
  \includegraphics[width=1\linewidth]{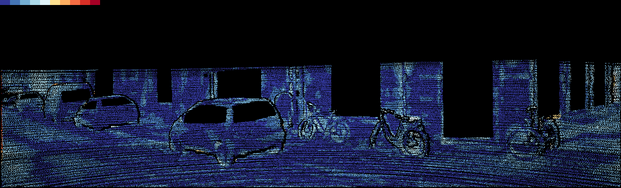}
\end{subfigure}

\caption{Generalization results on KITTI 2012 training set. The models are trained only on the SceneFlow dataset. The left panel shows the left input image and the ground truth disparity. For each example, the first row shows the colorized disparity prediction and the second row shows the D1 error map. In the D1 error map shows errors from lowest to highest using the color spectrum between blue and red.}
\label{2012fig}
\end{figure}

\begin{figure}
\centering
\begin{subfigure}{.22\textwidth}
  \centering
  \captionsetup{font=scriptsize, labelformat=empty} 
  \caption{Left image}
  \includegraphics[width=1\linewidth]{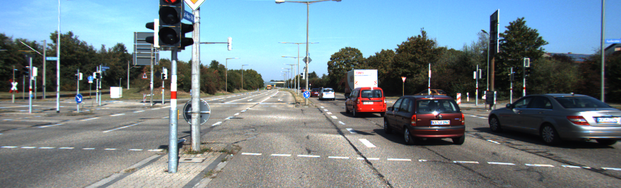}
\end{subfigure}
\begin{subfigure}{.22\textwidth}
  \centering
  \captionsetup{font=scriptsize, labelformat=empty} 
  \caption{ESMStereo-S-gwc}
  \includegraphics[width=1\linewidth]{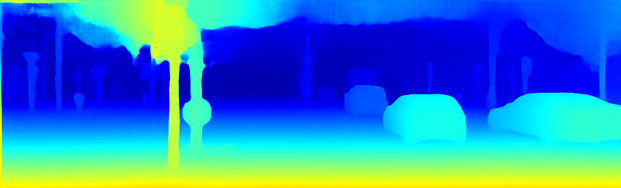}
\end{subfigure}
\begin{subfigure}{.22\textwidth}
  \centering
  \captionsetup{font=scriptsize, labelformat=empty} 
  \caption{ESMStereo-M-gwc}
  \includegraphics[width=1\linewidth]{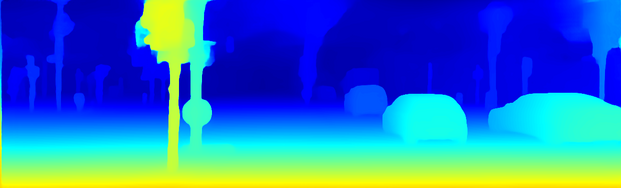}
\end{subfigure}
\begin{subfigure}{.22\textwidth}
  \centering
  \captionsetup{font=scriptsize, labelformat=empty} 
  \caption{ESMStereo-L-gwc}
  \includegraphics[width=1\linewidth]{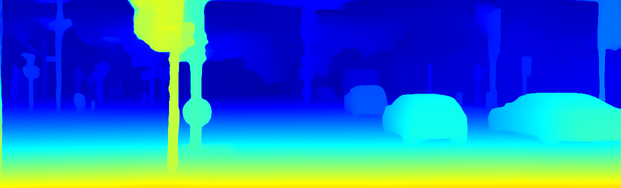}
\end{subfigure}
\hfill
\begin{subfigure}{.22\textwidth}
  \centering
  \includegraphics[width=1\linewidth]{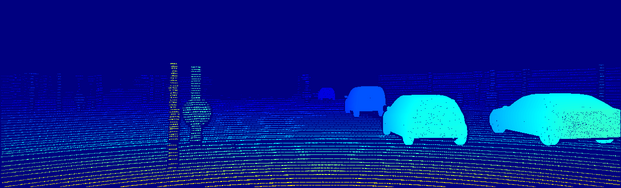}
\end{subfigure}
\begin{subfigure}{.22\textwidth}
  \centering
  \includegraphics[width=1\linewidth]{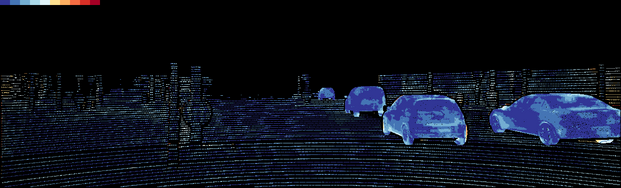}
\end{subfigure}
\begin{subfigure}{.22\textwidth}
  \centering
  \includegraphics[width=1\linewidth]{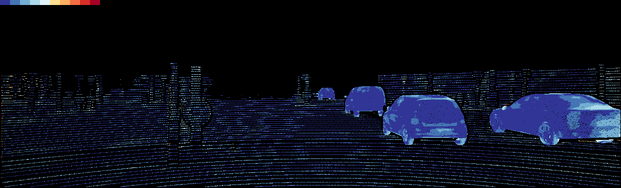}
\end{subfigure}
\begin{subfigure}{.22\textwidth}
  \centering
  \includegraphics[width=1\linewidth]{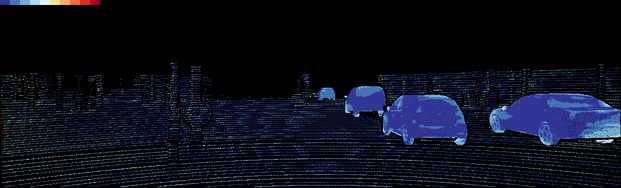}
\end{subfigure}
\begin{subfigure}{.22\textwidth}
  \centering

  \includegraphics[width=1\linewidth]{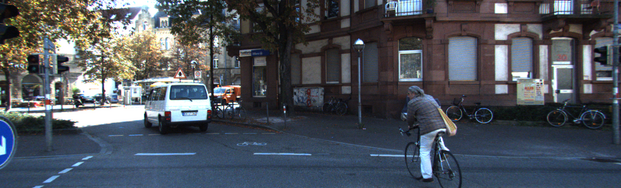}
\end{subfigure}
\begin{subfigure}{.22\textwidth}
  \centering
  \includegraphics[width=1\linewidth]{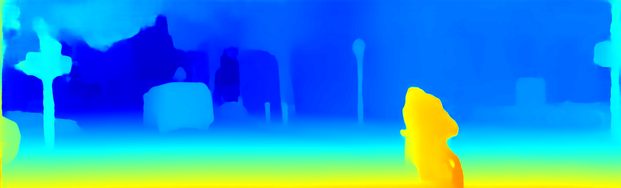}
\end{subfigure}
\begin{subfigure}{.22\textwidth}
  \centering
  \includegraphics[width=1\linewidth]{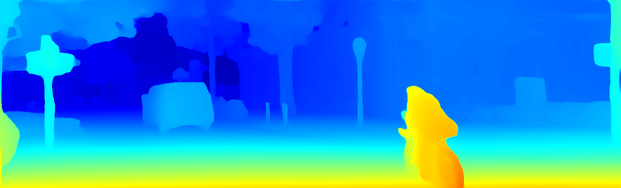}
\end{subfigure}
\begin{subfigure}{.22\textwidth}
  \centering
  \includegraphics[width=1\linewidth]{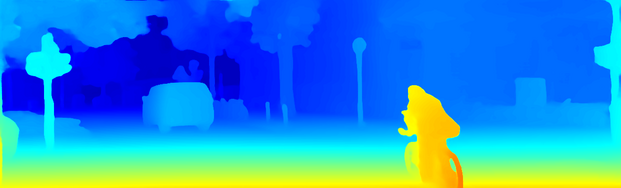}
\end{subfigure}
\hfill
\begin{subfigure}{.22\textwidth}
  \centering
  \includegraphics[width=1\linewidth]{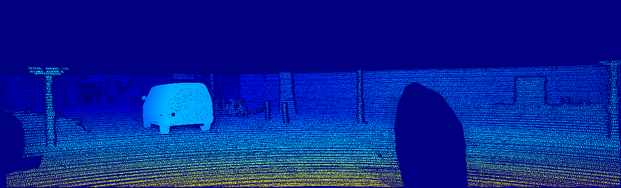}
\end{subfigure}
\begin{subfigure}{.22\textwidth}
  \centering
  \includegraphics[width=1\linewidth]{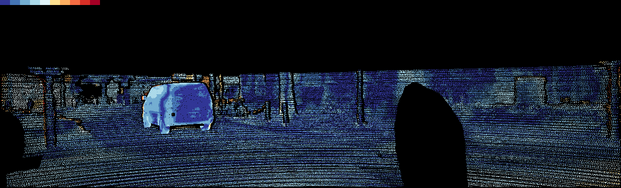}
\end{subfigure}
\begin{subfigure}{.22\textwidth}
  \centering
  \includegraphics[width=1\linewidth]{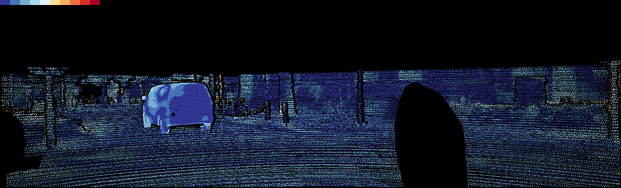}
\end{subfigure}
\begin{subfigure}{.22\textwidth}
  \centering
  \includegraphics[width=1\linewidth]{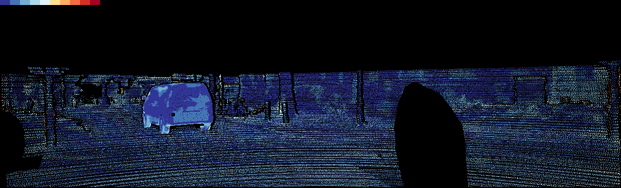}
\end{subfigure}
\begin{subfigure}{.22\textwidth}
  \centering

  \includegraphics[width=1\linewidth]{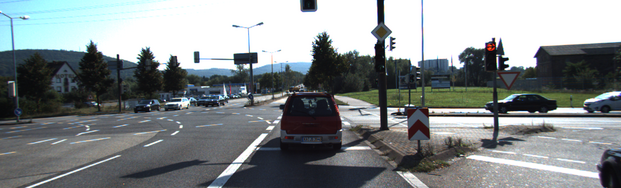}
\end{subfigure}
\begin{subfigure}{.22\textwidth}
  \centering
  \includegraphics[width=1\linewidth]{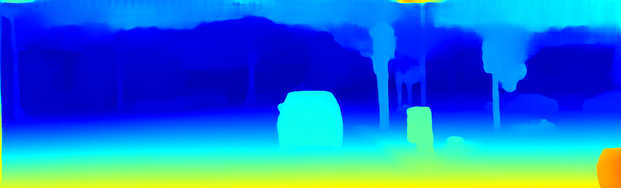}
\end{subfigure}
\begin{subfigure}{.22\textwidth}
  \centering
  \includegraphics[width=1\linewidth]{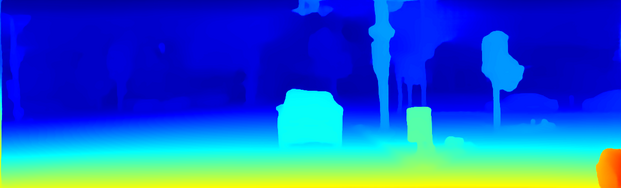}
\end{subfigure}
\begin{subfigure}{.22\textwidth}
  \centering
  \includegraphics[width=1\linewidth]{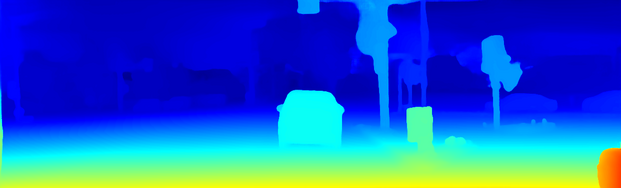}
\end{subfigure}
\hfill
\begin{subfigure}{.22\textwidth}
  \centering
  \includegraphics[width=1\linewidth]{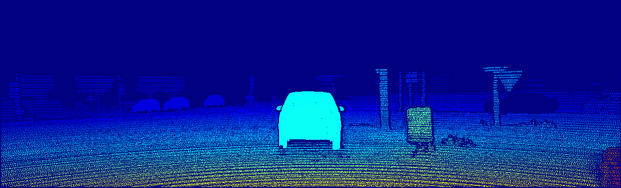}
\end{subfigure}
\begin{subfigure}{.22\textwidth}
  \centering
  \includegraphics[width=1\linewidth]{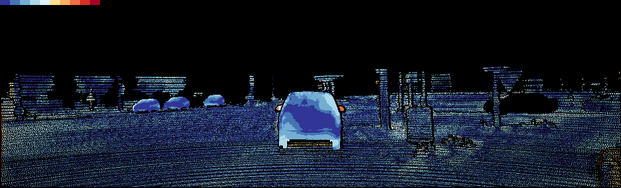}
\end{subfigure}
\begin{subfigure}{.22\textwidth}
  \centering
  \includegraphics[width=1\linewidth]{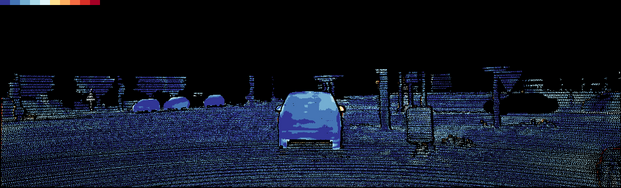}
\end{subfigure}
\begin{subfigure}{.22\textwidth}
  \centering
  \includegraphics[width=1\linewidth]{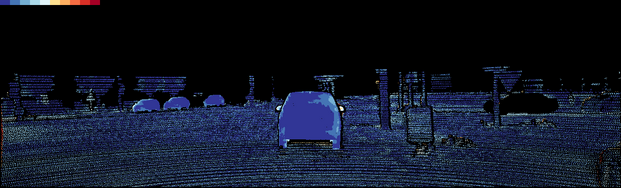}
\end{subfigure}
\caption{Generalization results on KITTI 2015 for the models only trained on SceneFlow dataset. The left panel shows the left input image and
the ground truth disparity. For each example, the first row shows the colorized disparity prediction and the second row shows the D1 error map. In the D1 error map lowest to highest error is demonstrated using the color spectrum between blue and red.}
\label{2015fig}
\end{figure}

\subsection{Generalization Performance}
\label{gen}

ESMStereo models are evaluated on the KITTI 2012 \cite{kitti_12}, KITTI 2015 \cite{kitti_15}, Middlebury 2014 \cite{midburry}, and ETH3D \cite{ETH3D} datasets to assess their generalization ability on real-world data. The models are trained exclusively on the SceneFlow dataset, and the results are reported in Table \ref{T4}. ESMStereo-L-gwc demonstrates strong generalization performance compared to real-time stereo matching methods which shows consistently competitive results across different real-world benchmarks. While it does not outperform all accuracy-oriented approaches on datasets such as Middlebury and ETH3D, it achieves a favorable trade-off between accuracy and efficiency that indicates robust adaptability across diverse scene types and imaging conditions. Furthermore, ESMStereo-S-gwc with just 8.5 ms inference time achieves a competitive generalization accuracy on KITTI 2012 and KITTI 2015. This strong generalization performance highlights the robustness and versatility of ESMStereo in predicting depth when an unseen scenario is presented.

\begin{table}[H]
\caption{Generalization performance on KITTI, Middlebury and
ETH3D. All models are trained only on SceneFlow. The bold font indicates best performance and underlined values indicate second best performance. The results for methods indicate by * are not reported in the original paper and computed as part of this research for benchmarking using their available source code.}
\centering
\begin{adjustbox}{width=1\textwidth}
\small
\begin{tabular}{c|l|c c c c}
\hline

\multirow{2}{*}{Target} & \multirow{2}{*}{Method} &  \multirow{2}{*}{\shortstack[c] {KITTI 2012 \\ D1[\%]}} &  \multirow{2}{*}{\shortstack[c] {KITTI 2015 \\ D1[\%]}} & \multirow{2}{*}{\shortstack[c] {Middlebury \\ bad 2.0[\%]}} & \multirow{2}{*}{\shortstack[c] {ETH3D \\ bad 1.0[\%]}} \\
& & & & &  \\\hline \hline

\parbox[t]{2mm}{\multirow{9}{*}{\rotatebox[origin=c]{90}{Accuracy}}} &PSMNet \cite{chang2018pyramid} &6.0 & 6.3&15.8 &9.8 \\ 
&GANet \cite{22_ganet}&10.1 & 11.7&20.3 &14.1  \\ 
&DSMNet \cite{zhang2020domain}& 6.2& 6.5& 13.8 & 6.2\\ 
&CFNet \cite{shen2021cfnet}&5.1 &6.0 &15.4 & 5.3 \\ 
&STTR \cite{li2021revisiting}&8.7 &6.7 &15.5 & 17.2 \\  
&FC-PSMNet \cite{zhang2022revisiting}&5.3 &5.8  &15.1 & 9.3\\ 
&Graft-PSMNet \cite{liu2022graftnet}&4.3 & 4.8 &9.7 &7.7 \\  
&IGEV-Stereo \cite{xu2023iterative}&- & - &6.2 &3.6 \\
&MoCha-Stereo \cite{chen2024mocha}&- & - &4.9 &3.2 \\\hline 
\parbox[t]{3mm}{\multirow{8}{*}{\rotatebox[origin=c]{90}{Speed (Real-Time)}}}  &DeepPrunerFast  \cite{18_deeppruner} &7.6 &7.6 &38.7 &36.8\\ 
&BGNet \cite{9_bgnet} &12.5 &11.7 &24.7 &22.6\\ 
&CoEx \cite{12_coex} &7.6 &7.2 &14.5 &9.0\\ 
&CGI-Stereo \cite{11_cgistereo}&6.0 &\underline{5.8} &13.5 &6.3\\ 
&LightStereo-M \cite{guo2024lightstereo}* &16.4 &17.7 & - & -\\
&LightStereo-L \cite{guo2024lightstereo}* &12.1 &15.2 & - & -\\
& IINet \cite{li2024iinet}  &11.6 & 8.5& 19.5 & -\\
& Fast-ACVNet \cite{25_acvnet}  &12.4 & 10.6& 20.1 & -\\\hline
& \textbf{ESMStereo-S-gwc}  &12.3 &10.8 &24.1 & 27.8 \\
& \textbf{ESMStereo-M-gwc}  &9.7 &8.2 &15.7 & 11.7 \\
& \textbf{ESMStereo-L-gwc}  &\textbf{5.4} & \textbf{5.5} &\textbf{11.1} & \textbf{6.1} \\ \hline
\end{tabular}
\end{adjustbox}

\label{T4}
\end{table}

{\setlength{\parindent}{0cm}
\vspace{5mm}
\par
Although ESMStereo-L-gwc demonstrates strong overall generalization among real-time stereo matching methods, heavier models such as IGEV-Stereo \cite{xu2023iterative} and MoCha-Stereo \cite{chen2024mocha} achieve better performance on challenging ETH3D metrics (e.g., Bad 1.0), largely due to their substantially higher model capacity and more expensive iterative refinement strategies. These methods rely on deeper cost aggregation and repeated update mechanisms that are particularly effective in challenging scenarios such as large textureless regions, thin structures, and severe occlusions, where accurate disparity estimation requires extensive context aggregation and long-range reasoning. In contrast, ESMStereo-L-gwc is designed for real-time efficiency using compact cost volumes and lightweight refinement modules. While this design enables significantly faster inference, it inevitably limits the amount of global context modeling and fine-grained disparity correction, especially in ambiguous regions with weak or repetitive textures. Additionally, the reduced-resolution disparity representation used for efficiency further constrains the amount of high-frequency geometric detail that can be preserved during reconstruction.
{\setlength{\parindent}{0cm}
\vspace{5mm}
\par

Figs.\ref{2012fig} and \ref{2015fig} present qualitative results for the KITTI 2012 and KITTI 2015 training sets across all ESMStereo models. The D1 error maps are visualized using a color spectrum ranging from blue (lowest error) to red (highest error), illustrating the models' ability to recover thin and smooth structures. Similarly, qualitative results on the Middlebury and ETH3D datasets, shown in Figs.\ref{midfig} and \ref{ethfig}, indicate that all models effectively recover fine and thin structures. 

\begin{figure}[H]
\centering
\begin{subfigure}{.4\textwidth}
  \centering
     \caption{Left Image}
  \includegraphics[width=1\linewidth]{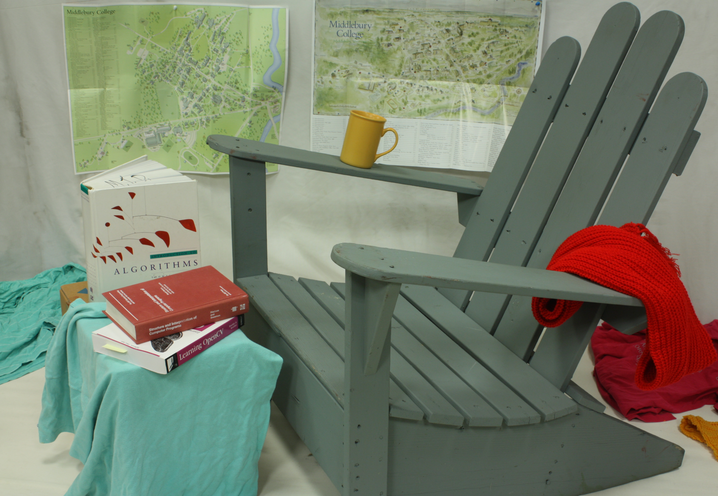}%
  \label{fig1:submid}
\end{subfigure}
\begin{subfigure}{.4\textwidth}
  \centering
    \caption{Ground truth}
  \includegraphics[width=1\linewidth]{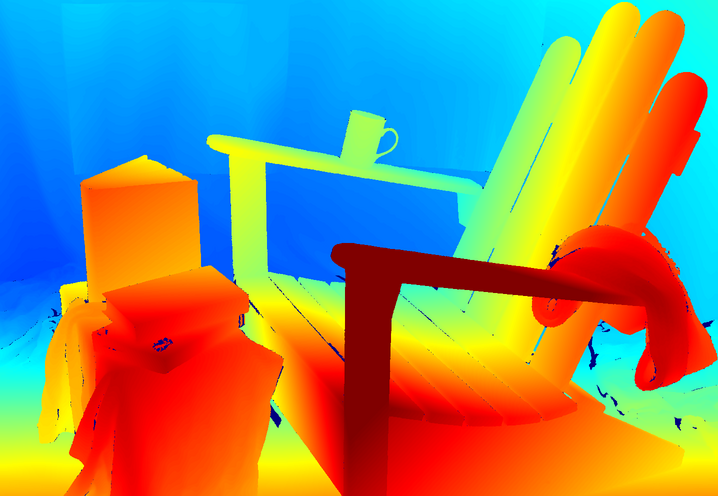}%
  \label{fig3:submid}
\end{subfigure}
\hfill
\begin{subfigure}{.3\textwidth}
  \centering
    \caption{ESMStereo-S-gwc}
  \includegraphics[width=1\linewidth]{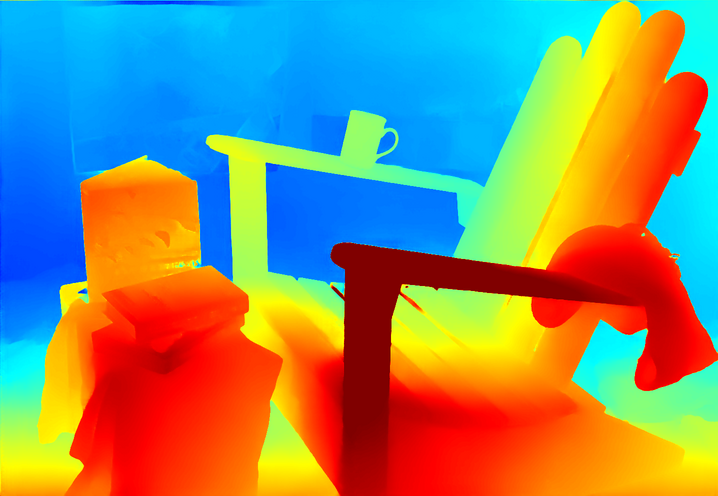}%
  \label{fig28:submidg}
\end{subfigure}
\begin{subfigure}{.3\textwidth}
  \centering
    \caption{ESMStereo-M-gwc}
  \includegraphics[width=1\linewidth]{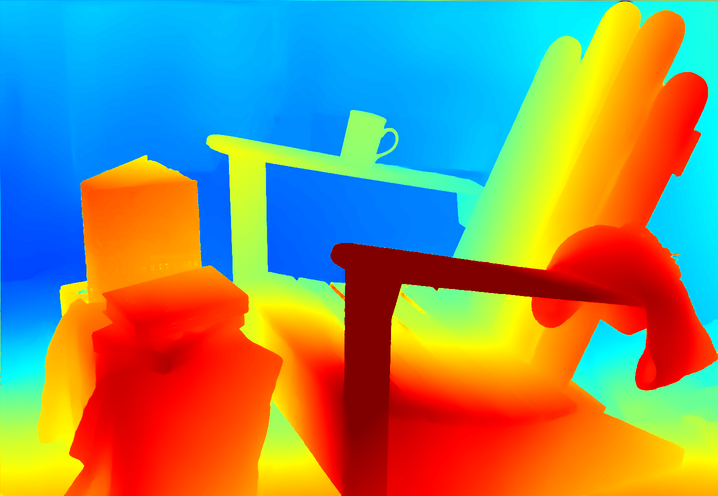}%
  \label{fig24:submidg}
\end{subfigure}
\begin{subfigure}{.3\textwidth}
  \centering
    \caption{ESMStereo-L-gwc}
  \includegraphics[width=1\linewidth]{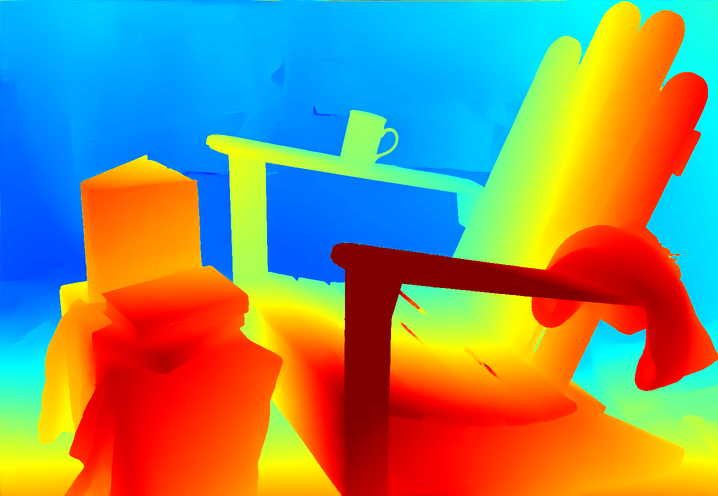}%
  \label{fig26:submidg}
\end{subfigure}
\hfill%
\begin{subfigure}{.3\textwidth}
  \centering
  \includegraphics[width=1\linewidth]{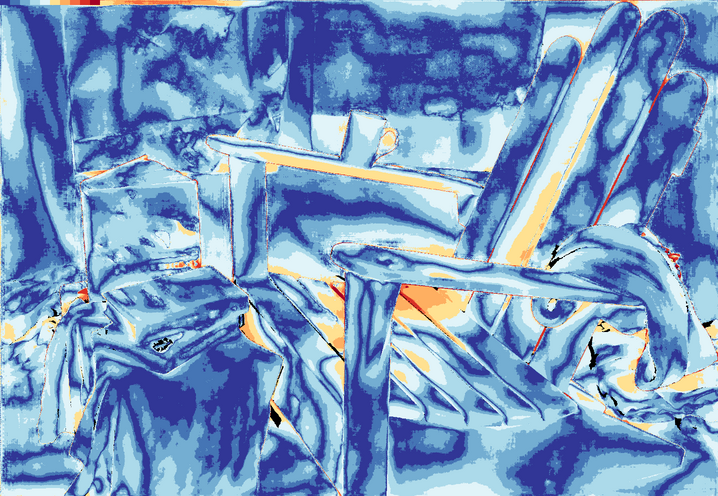}
  \label{fig28:submid}
\end{subfigure}
\begin{subfigure}{.3\textwidth}
  \centering
  \includegraphics[width=1\linewidth]{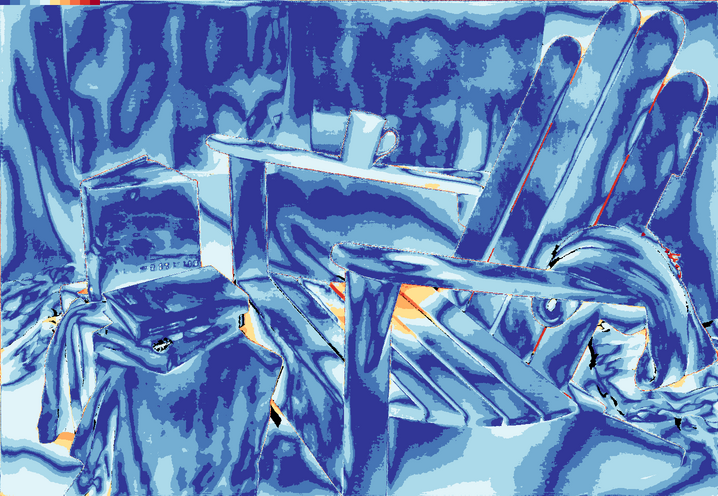}
  \label{fig24:submid}
\end{subfigure}
\begin{subfigure}{.3\textwidth}
  \centering
  \includegraphics[width=1\linewidth]{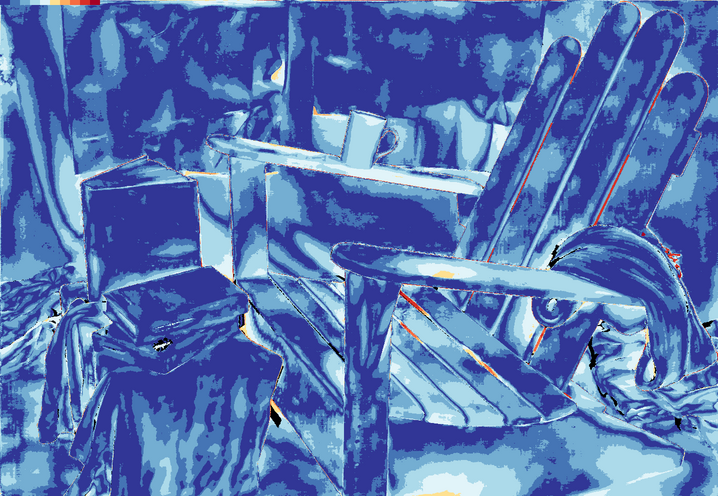}
  \label{fig26:submid}
\end{subfigure}

\caption{Generalization results of ESMStereo on Middlebury 2014 dataset when only trained on the synthetic SceneFlow dataset. The first row shows the left image and the ground truth, the second row represents the colorized disparity prediction and the third row shows the D1 error map.}
\label{midfig}
\end{figure}

\begin{figure}[H]
\centering
\begin{subfigure}{.4\textwidth}
  \centering
    \caption{Left image}
  \includegraphics[width=1\linewidth]{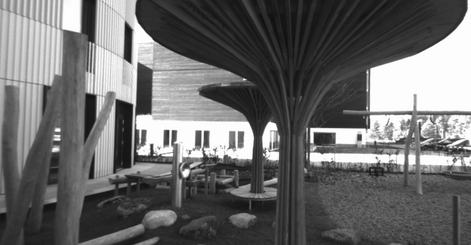}%
  \label{fig1:subeth}
\end{subfigure}
\begin{subfigure}{.4\textwidth}
  \centering
    \caption{Ground truth}
  \includegraphics[width=1\linewidth]{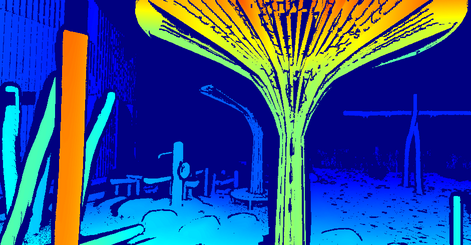}%
  \label{fig3:subeth}
\end{subfigure}
\begin{subfigure}{.3\textwidth}
  \centering
    \caption{ESMStereo-S-gwc}
  \includegraphics[width=1\linewidth]{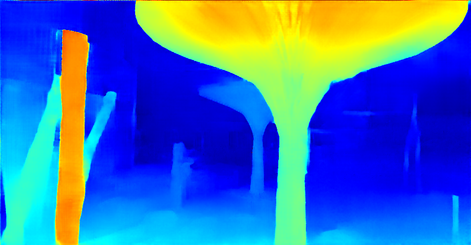}%
  \label{fig2:subethg}
\end{subfigure}
\begin{subfigure}{.3\textwidth}
  \centering
    \caption{ESMStereo-M-gwc}
  \includegraphics[width=1\linewidth]{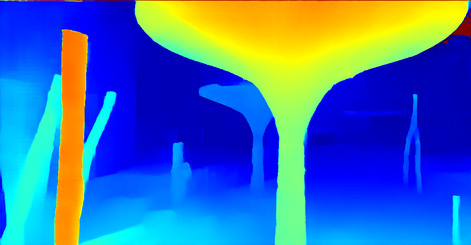}%
  \label{fig2:subeth4g}
\end{subfigure}
\begin{subfigure}{.3\textwidth}
  \centering
    \caption{ESMStereo-L-gwc}
  \includegraphics[width=1\linewidth]{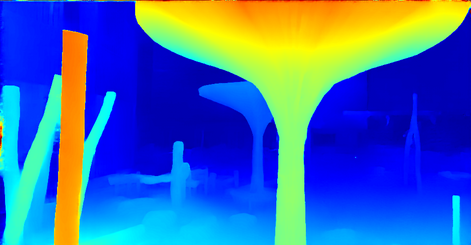}%
  \label{fig2:subeth}
\end{subfigure}
\hfill
\begin{subfigure}{.3\textwidth}
  \centering
  \includegraphics[width=1\linewidth]{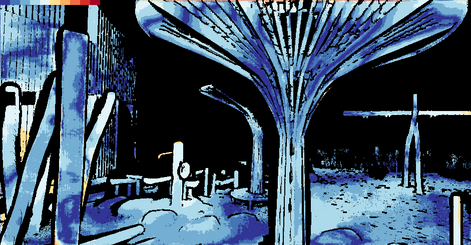}
  \label{fig28:subeth4}
\end{subfigure}
\begin{subfigure}{.3\textwidth}
  \centering
  \includegraphics[width=1\linewidth]{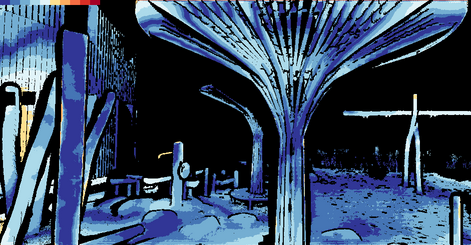}
  \label{fig26:subeth4}
\end{subfigure}
\begin{subfigure}{.3\textwidth}
  \centering
  \includegraphics[width=1\linewidth]{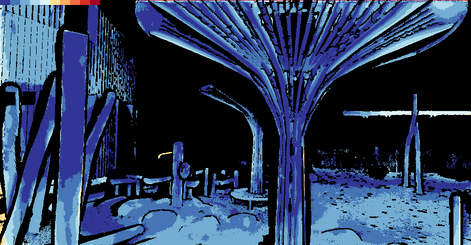}
  \label{fig25:subeth4}
\end{subfigure}
\caption{Generalization results of ESMStereo on ETH3D dataset. The first row shows the left image and the ground truth, the second row represents the colorized disparity prediction and the third row shows the D1 error map.}
\label{ethfig}
\end{figure}

\subsection{ESM Portability}

To evaluate the portability of the proposed ESM module, it is integrated into PSMNet \cite{chang2018pyramid} and Fast-ACVNet-Plus \cite{25_acvnet}, and the modified networks are retrained on the SceneFlow dataset. As shown in Table \ref{portable}, incorporating ESM consistently improves disparity estimation accuracy for both architectures, while introducing only a moderate increase in computational cost and parameter count. This increase is partly attributed to the heavier feature extraction backbones and aggregation units used in these architectures compared to the lightweight designs adopted in ESMStereo. In addition, qualitative results on the KITTI 2015 dataset demonstrate that the ESM-enhanced models produce smoother disparity maps and recover finer scene structures. These experiments primarily aim to validate that the proposed ESM module can be effectively integrated into different stereo matching backbones and consistently improve their feature refinement capability. It should be noted that the ESM module was not specifically re-tuned for runtime optimization within these external architectures, and therefore the reported results are intended to demonstrate portability and effectiveness rather than optimized deployment efficiency.

\begin{table}
\centering
\caption{ESM integration with other stereo matching methods in terms of accuracy and computational cost.}
\label{portable}

\begin{adjustbox}{width=0.7\textwidth}
\small
\begin{tabular}{l|c|c|c}
\hline
Architecture & EPE [px] & FLOPs [G] & Params [M]  \\
\hline
PSMNet \cite{chang2018pyramid}  & 1.09 & 613 & 5.22 \\ 
PSMNet + ESM & \textbf{1.02} & 756 & 5.70 \\ 
\hline
Fast-ACVNet-Plus \cite{25_acvnet} & 0.59 & 62 & 3.2  \\
Fast-ACVNet-Plus + ESM & \textbf{0.51} & 70 & 3.5  \\
\hline
\end{tabular}
\end{adjustbox}

\end{table}

\begin{figure}
\centering
\begin{subfigure}{.15\textwidth}
  \centering
  \captionsetup{font=scriptsize, labelformat=empty} 
  \caption{Left image}
  \includegraphics[width=1\linewidth]{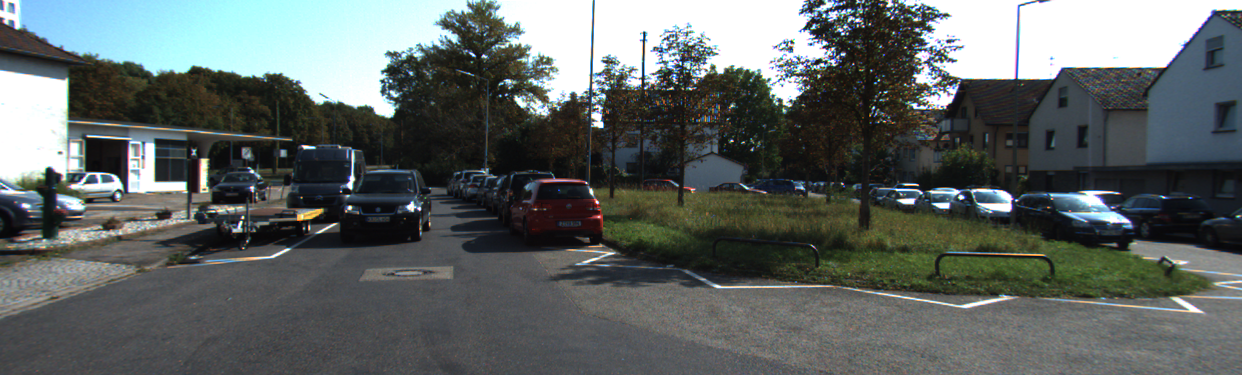}
\end{subfigure}
\begin{subfigure}{.15\textwidth}
  \centering
  \captionsetup{font=scriptsize, labelformat=empty, justification=centering} 
  \caption{PSMNet \cite{chang2018pyramid}}
  \includegraphics[width=1\linewidth]{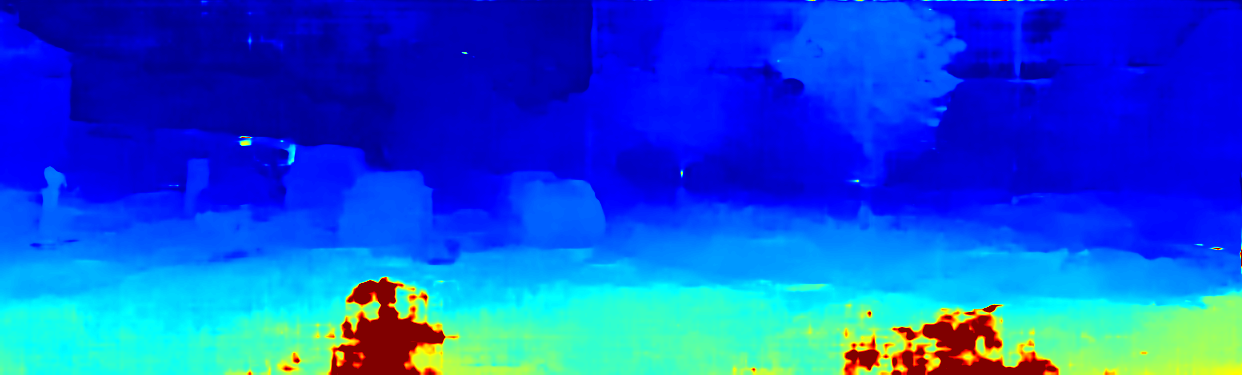}
\end{subfigure}
\begin{subfigure}{.15\textwidth}
  \centering
  \captionsetup{font=scriptsize, labelformat=empty, justification=centering} 
  \caption{PSMNet+ESM}
  \includegraphics[width=1\linewidth]{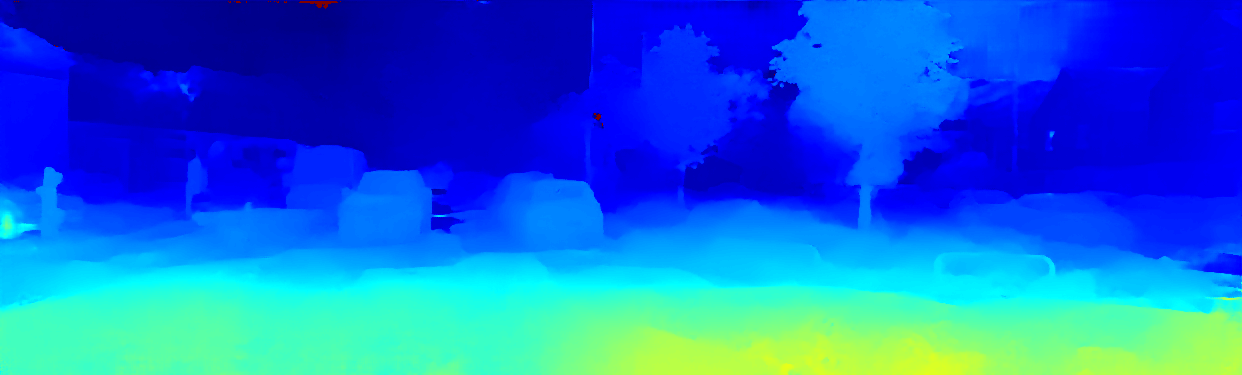}
\end{subfigure}
\begin{subfigure}{.15\textwidth}
  \centering
  \captionsetup{font=scriptsize, labelformat=empty} 
  \caption{Left image}
  \includegraphics[width=1\linewidth]{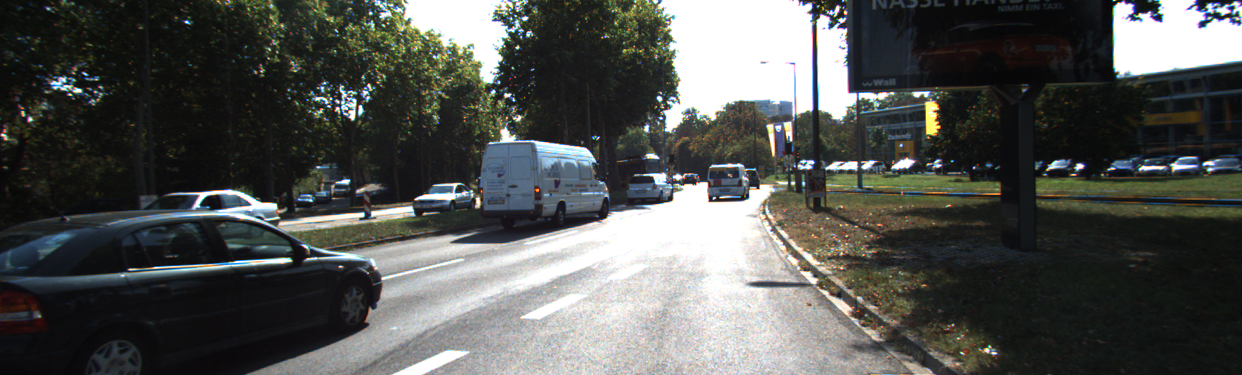}
\end{subfigure}
\begin{subfigure}{.15\textwidth}
  \centering
  \captionsetup{font=scriptsize, labelformat=empty, justification=centering} 
  \caption{Fast-ACVNet-Plus \cite{25_acvnet}}
  \includegraphics[width=1\linewidth]{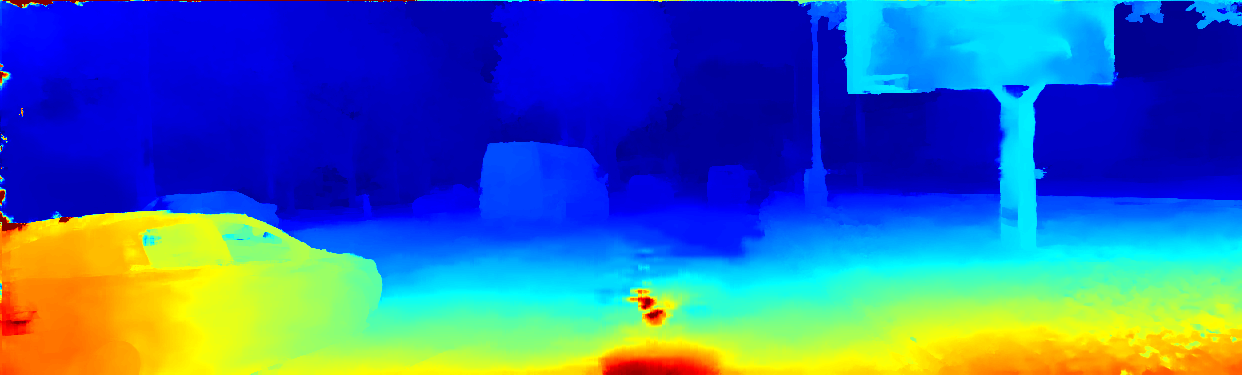}
\end{subfigure}
\begin{subfigure}{.15\textwidth}
  \centering
  \captionsetup{font=scriptsize, labelformat=empty, justification=centering} 
  \caption{Fast-ACVNet-Plus+ESM}
  \includegraphics[width=1\linewidth]{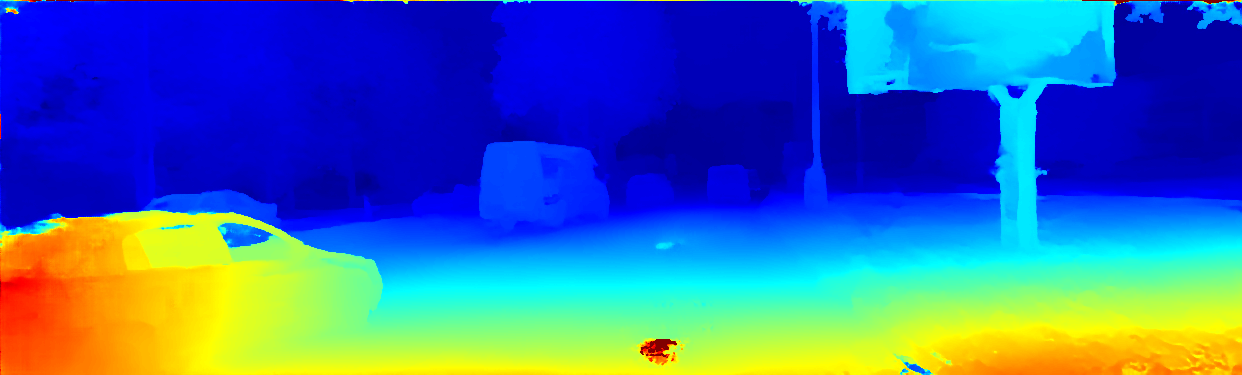}
\end{subfigure}
\hfill
\begin{subfigure}{.15\textwidth}
  \centering
  \captionsetup{font=scriptsize, labelformat=empty, justification=centering} 
  \includegraphics[width=1\linewidth]{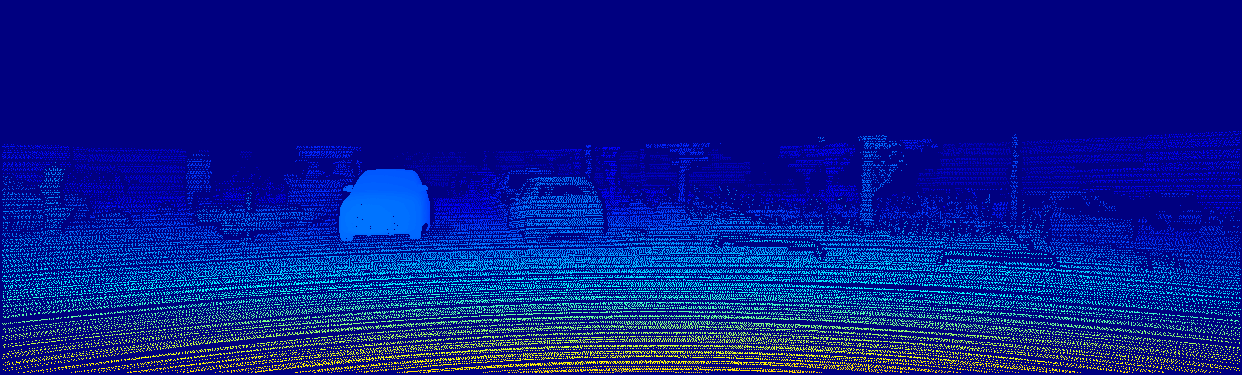}
\end{subfigure}
\begin{subfigure}{.15\textwidth}
  \centering
  \captionsetup{font=scriptsize, labelformat=empty, justification=centering} 
  \includegraphics[width=1\linewidth]{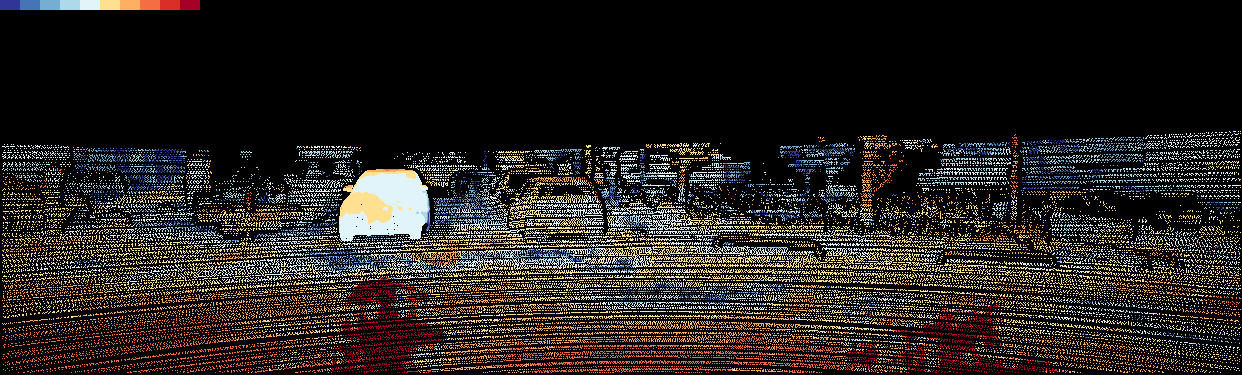}
\end{subfigure}
\begin{subfigure}{.15\textwidth}
  \centering
  \captionsetup{font=scriptsize, labelformat=empty, justification=centering} 
  \includegraphics[width=1\linewidth]{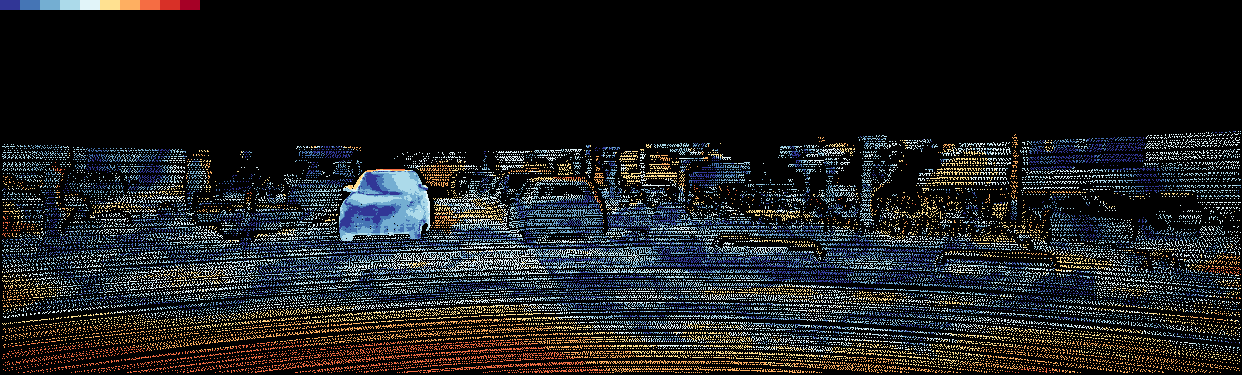}
\end{subfigure}
\begin{subfigure}{.15\textwidth}
  \centering
  \captionsetup{font=scriptsize, labelformat=empty} 
  \includegraphics[width=1\linewidth]{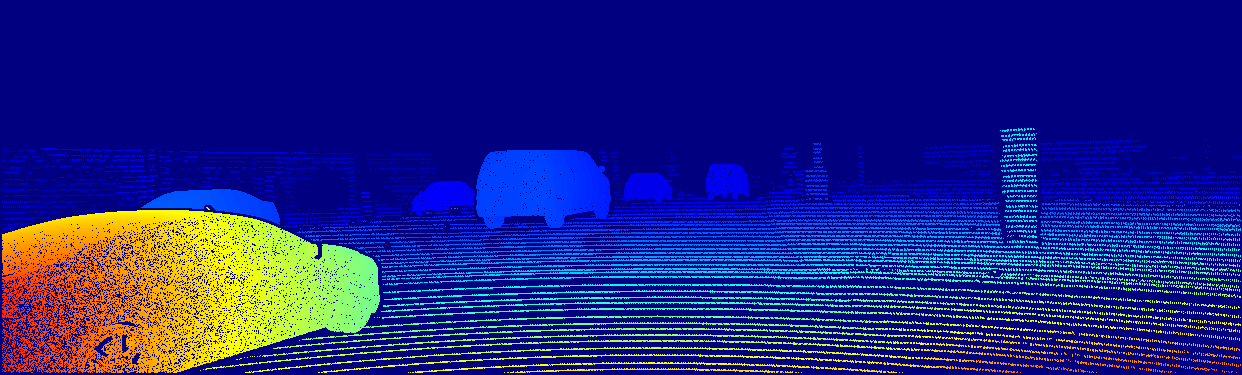}
\end{subfigure}
\begin{subfigure}{.15\textwidth}
  \centering
  \captionsetup{font=scriptsize, labelformat=empty, justification=centering} 
  \includegraphics[width=1\linewidth]{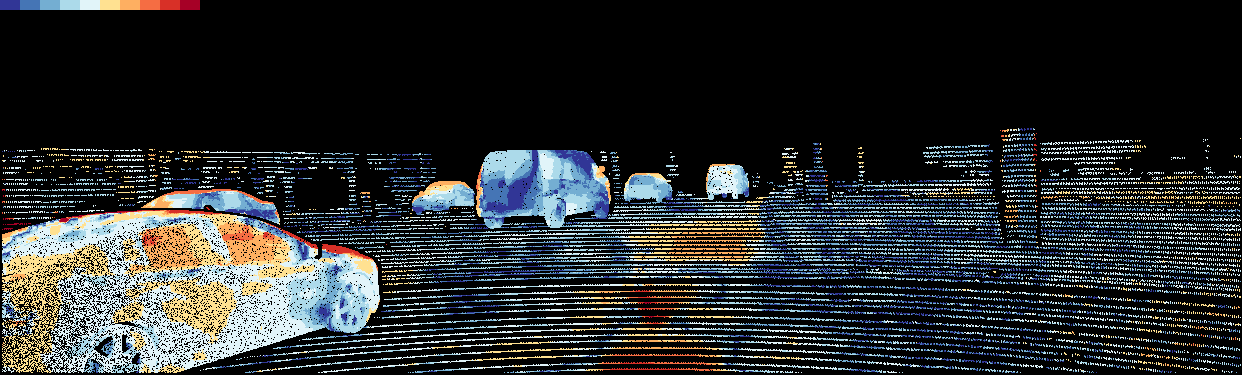}
\end{subfigure}
\begin{subfigure}{.15\textwidth}
  \centering
  \captionsetup{font=scriptsize, labelformat=empty, justification=centering} 
  \includegraphics[width=1\linewidth]{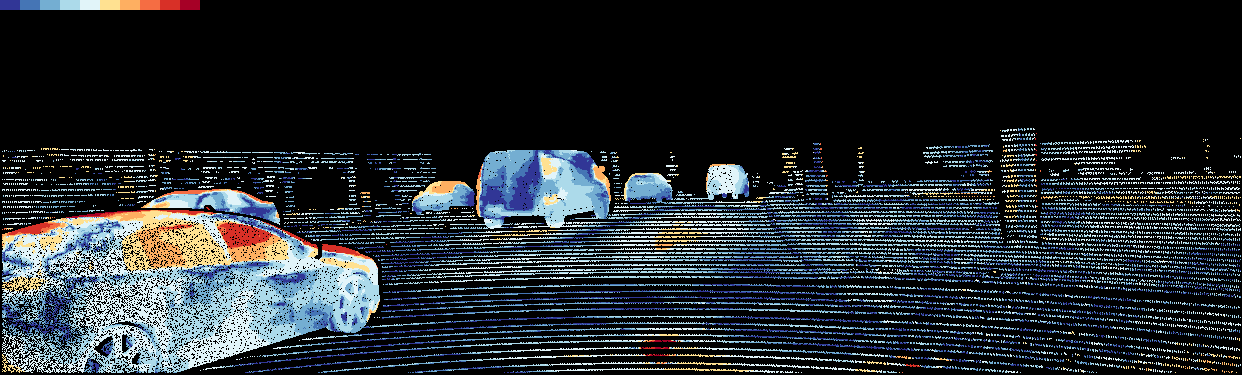}
\end{subfigure}
\hfill
\begin{subfigure}{.15\textwidth}
  \centering
  \includegraphics[width=1\linewidth]{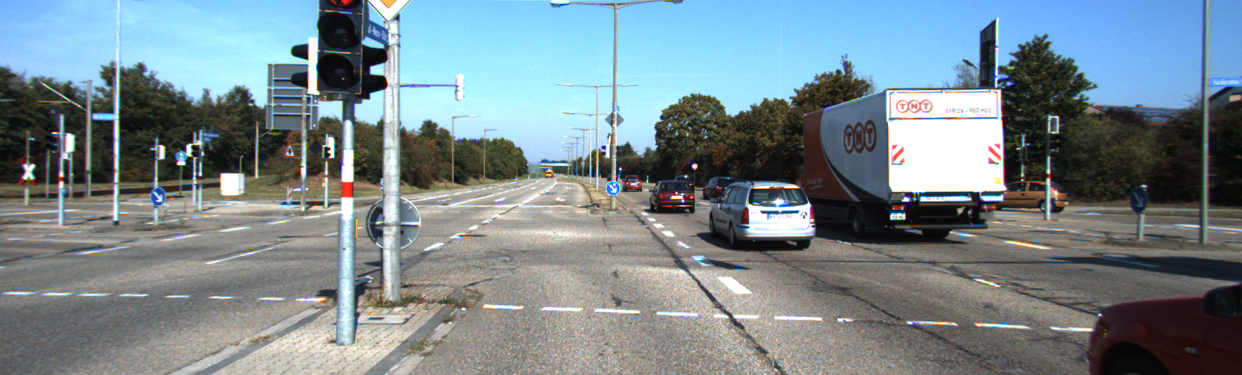}
\end{subfigure}
\begin{subfigure}{.15\textwidth}
  \centering
  \includegraphics[width=1\linewidth]{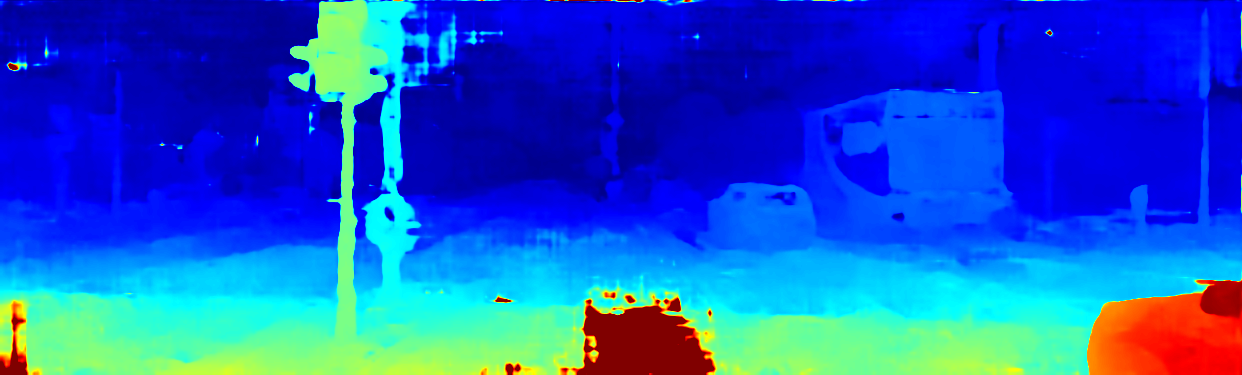}
\end{subfigure}
\begin{subfigure}{.15\textwidth}
  \centering
  \includegraphics[width=1\linewidth]{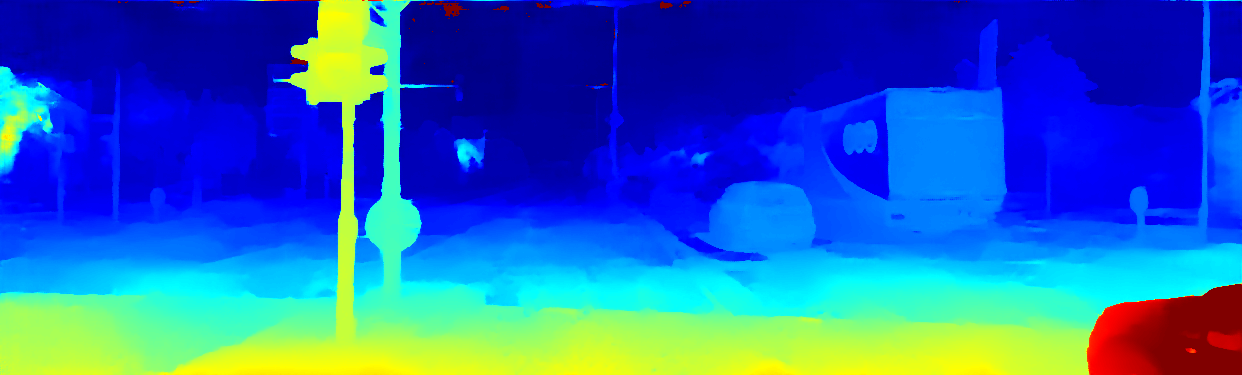}
\end{subfigure}
\begin{subfigure}{.15\textwidth}
  \centering
  \includegraphics[width=1\linewidth]{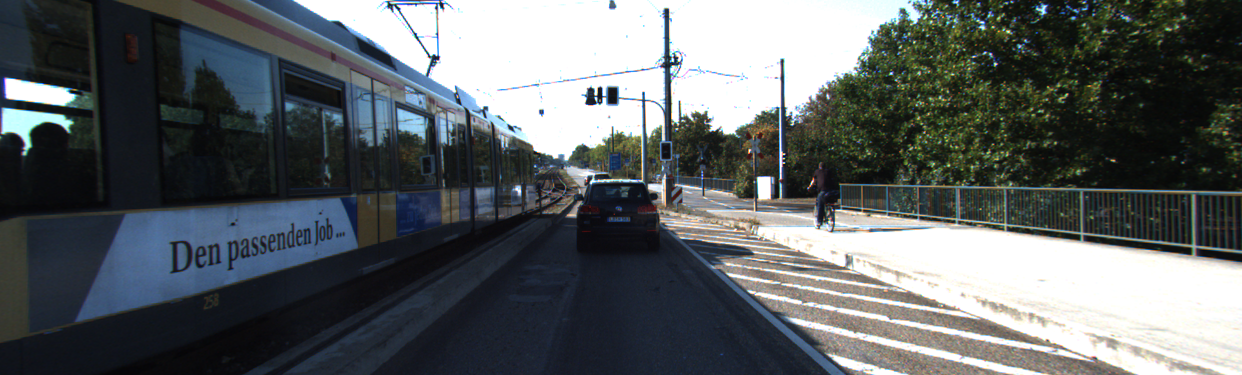}
\end{subfigure}
\begin{subfigure}{.15\textwidth}
  \centering
  \includegraphics[width=1\linewidth]{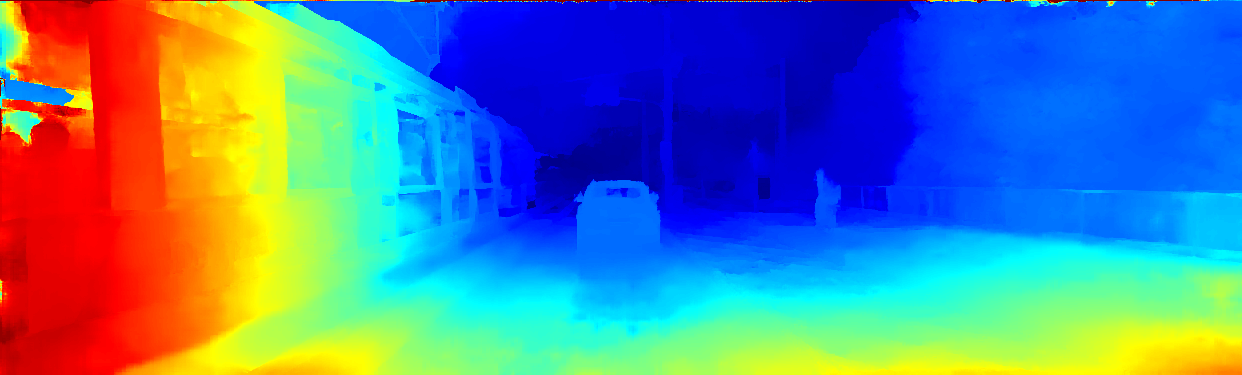}
\end{subfigure}
\begin{subfigure}{.15\textwidth}
  \centering
  \includegraphics[width=1\linewidth]{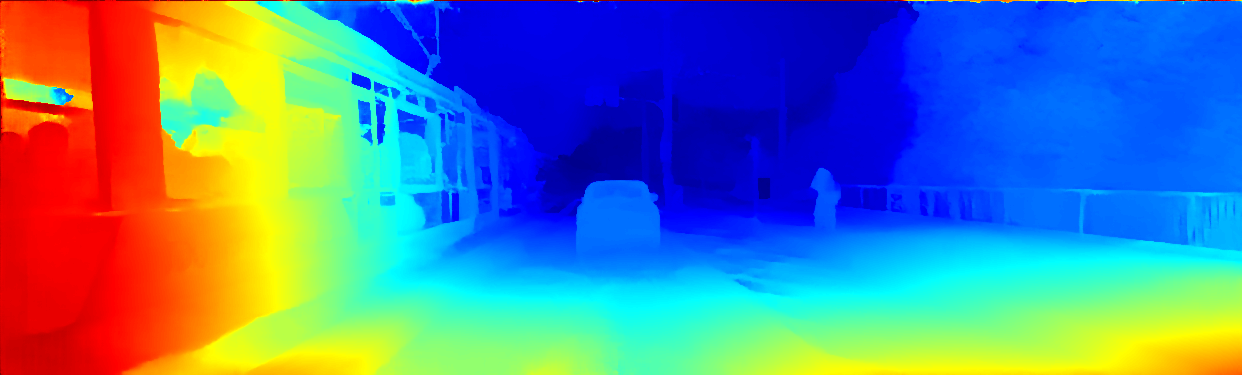}
\end{subfigure}
\hfill
\begin{subfigure}{.15\textwidth}
  \centering
  \includegraphics[width=1\linewidth]{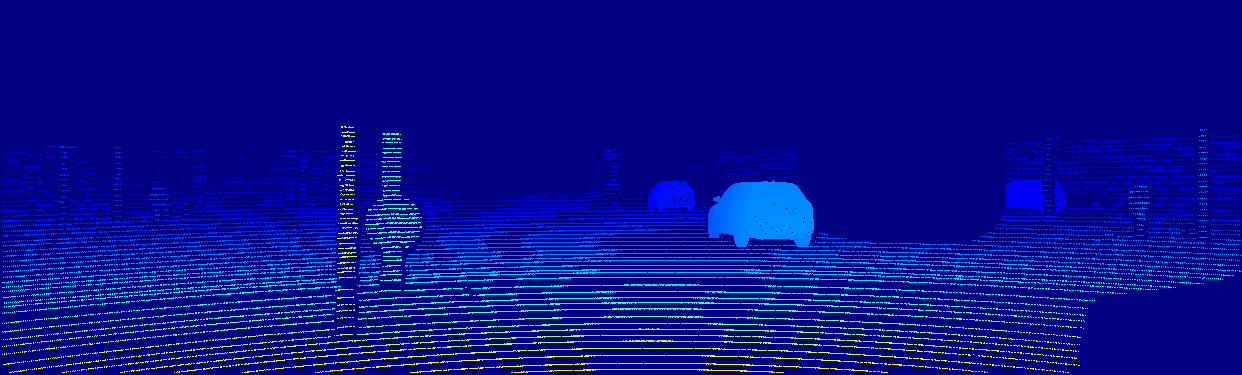}
\end{subfigure}
\begin{subfigure}{.15\textwidth}
  \centering
  \includegraphics[width=1\linewidth]{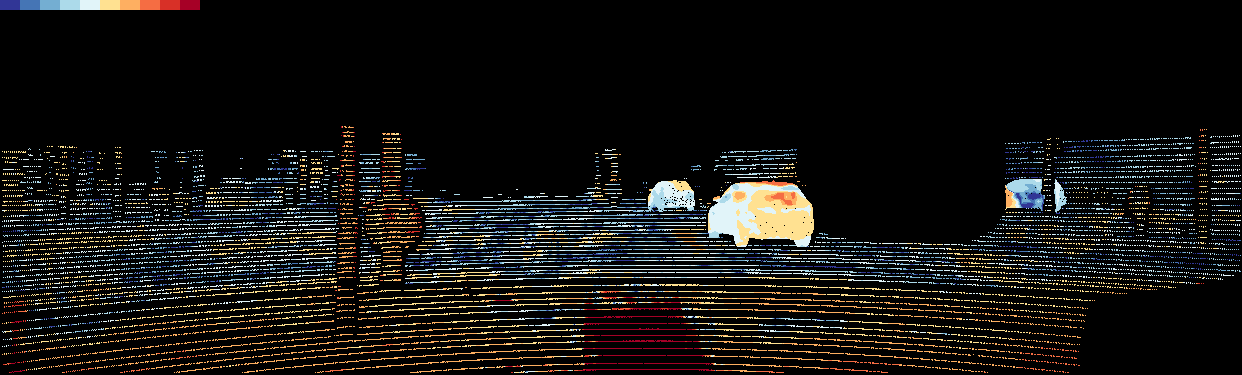}
\end{subfigure}
\begin{subfigure}{.15\textwidth}
  \centering
  \includegraphics[width=1\linewidth]{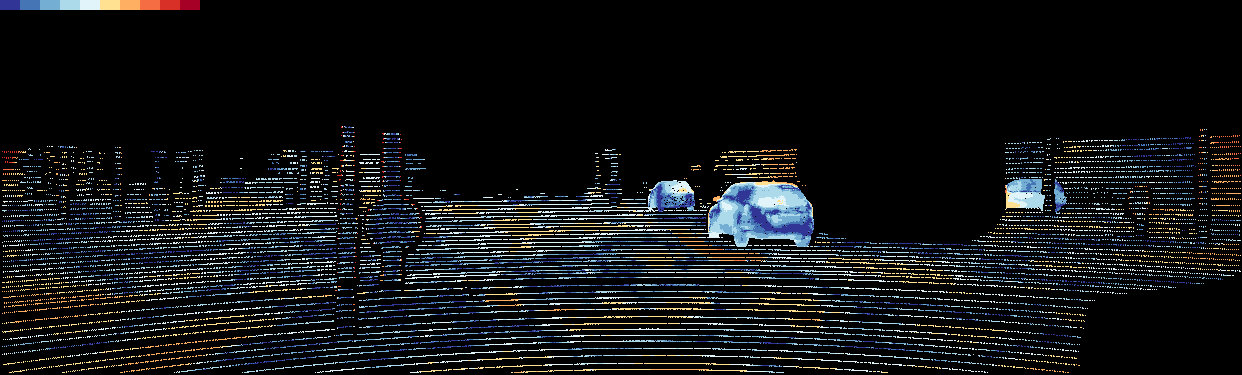}
\end{subfigure}
\begin{subfigure}{.15\textwidth}
  \centering
  \includegraphics[width=1\linewidth]{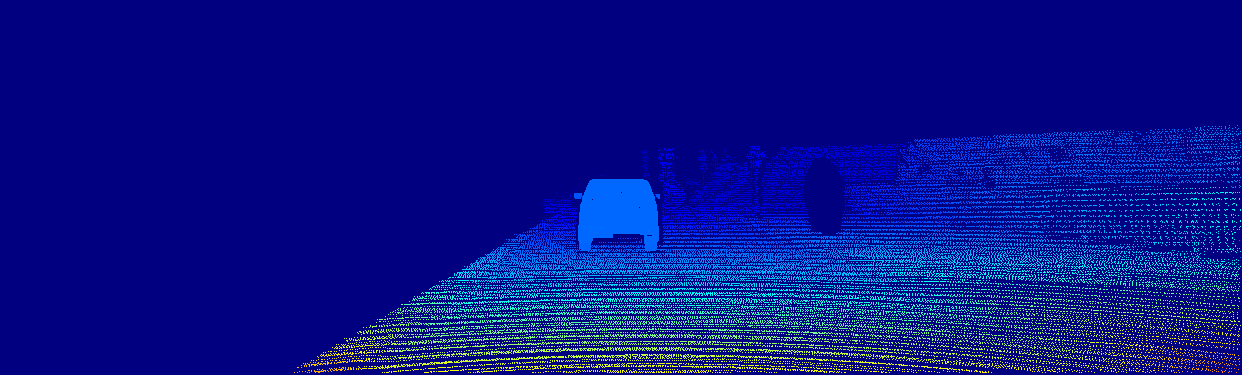}
\end{subfigure}
\begin{subfigure}{.15\textwidth}
  \centering
  \includegraphics[width=1\linewidth]{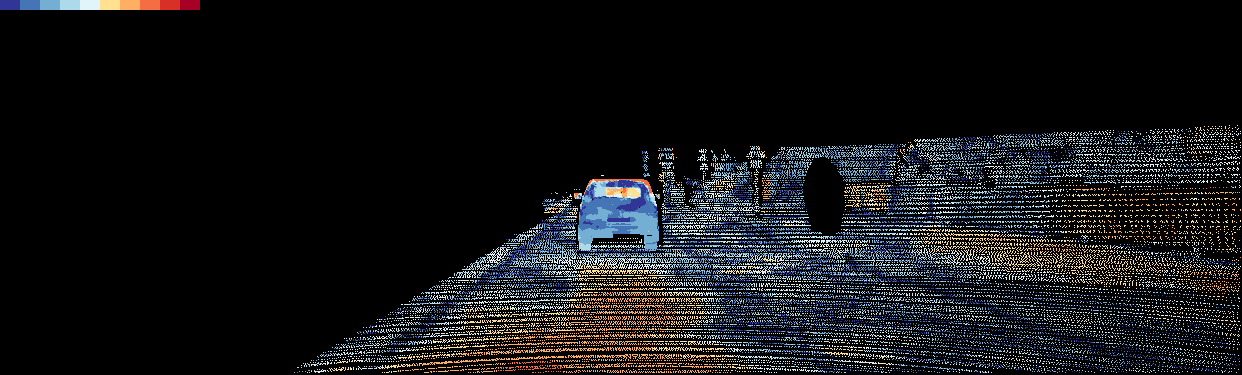}
\end{subfigure}
\begin{subfigure}{.15\textwidth}
  \centering
  \includegraphics[width=1\linewidth]{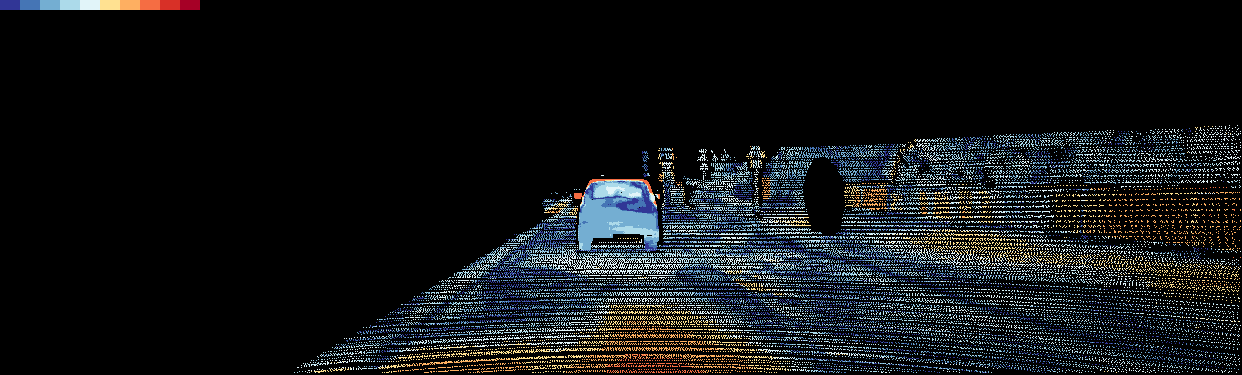}
\end{subfigure}
\caption{Qualitative results on KITTI 2015 for the integrated ESM module. The left panel shows the left input image and the ground truth disparity, the second column is the results without ESM and the third column is the results with ESM . For each example, the first row shows the colorized disparity prediction and the second row shows the D1 error map.}
\label{gen_comp}
\end{figure}

\subsection{Computational Complexity Analysis}
The low computational complexity of the ESMStereo models stems from reducing the computational demands of their 3D components, specifically the cost volumes and aggregation blocks. By employing compact cost volumes and lightweight aggregation blocks, and shifting the computational burden to post-disparity regression, the models significantly reduce Floating Point Operations per Second (FLOPs). This is achieved by using 2D ESM modules to recover information that may be lost due to the simplified cost volumes and aggregation units.

{\setlength{\parindent}{0cm}
\vspace{5mm}
\par

For instance, ESMStereo-S-gwc achieves an End-Point Error (EPE) of 1.10 pixel on the SceneFlow dataset with just 1.7 million parameters with a 91 FPS inference time on Jetson AGX. Meanwhile, ESMStereo-L-gwc delivers an even more impressive accuracy, achieving an EPE of 0.53 pixel with only 6.6 million parameters. In comparison, LightStereo-H \cite{guo2024lightstereo} attains EPE of 0.51 pixels but requires a significantly larger model with 45.6 million parameters. Therefore, the combination of low computational complexity and high accuracy makes ESMStereo highly suitable for real-time deployment on edge devices.

\section{Conclusion}

The ESMStereo models are designed with low computational complexity to achieve an acceptable balance between accuracy and speed, which makes them ideal for real-time stereo matching with exceptional performance on real-world datasets. These pipelines excel in capturing thin structures and managing complex visual environments effectively. The fast inference speed and high accuracy of ESMStereo are driven by two core design considerations. First is the use of a compact cost volume, processed by a lightweight 3D aggregation module, and second is the Enhanced ShuffleMixer (ESM) module. The ESM module functions as a unified upsampling component, enhancing accuracy by integrating the disparity map with image features and efficient feature mixing. This design enables the network to capture both global and local contexts to recover the information loss associated with compact cost volumes and lightweight aggregation blocks and obtain highly accurate disparity maps in real-time. As shown in the experimental results, ESMStereo-L-gwc outperforms state-of-the-art real-time methods in terms of accuracy and generalization ability.

\pagebreak
\vspace{5mm}
\par
\bibliography{esmstereo}
\end{document}